%% file: arxiv_recyclablegp.tex
\documentclass{article}

\usepackage[]{arxiv}

\usepackage{natbib}
\setcitestyle{authoryear,open={(},close={)}}

\usepackage[utf8]{inputenc} % allow utf-8 input
\usepackage[T1]{fontenc}    % use 8-bit T1 fonts
\usepackage{hyperref}       % hyperlinks
\usepackage{url}            % simple URL typesetting
\usepackage{booktabs}       % professional-quality tables
\usepackage{amsfonts}       % blackboard math symbols
\usepackage{nicefrac}       % compact symbols for 1/2, etc.
\usepackage{microtype}      % microtypography
\usepackage{bm}
\usepackage{amsmath}
\usepackage{amsfonts}
\usepackage{color}
\usepackage[makeroom]{cancel}
\usepackage{cancel}
\usepackage{tabu}
\usepackage[normalem]{ulem} %to strike the words
\usepackage{framed}
\usepackage[pdftex]{graphicx}
\usepackage{subcaption}
\usepackage{wrapfig}
\usepackage{amssymb}
\usepackage{pifont}
\usepackage{multirow}
\usepackage{booktabs}
\usepackage[flushleft]{threeparttable}
\usepackage{float}
\usepackage{algorithm}
\usepackage{algorithmic}

\input{definitions}
\title{Recyclable Gaussian Processes}

\author{
	Pablo Moreno-Muñoz\\%\hspace*{2.0cm}Antonio Art\'es-Rodr\'iguez\\ 
	Dept. of Signal Theory and Communications\\
	Universidad Carlos III de Madrid, Spain\\
	\texttt{pmoreno@tsc.uc3m.es} \\
	\And
	Antonio Art\'es-Rodr\'iguez\\
	Dept. of Signal Theory and Communications\\
	Universidad Carlos III de Madrid, Spain\\
	\texttt{antonio@tsc.uc3m.es} \\
	\And
	Mauricio A. \'Alvarez\\ 
	Dept. of Computer Science\\
	University of Sheffield, UK\\
	\texttt{mauricio.alvarez@sheffield.ac.uk} \\
}
\begin{document}

\maketitle

\begin{abstract}
	We present a new framework for recycling independent variational approximations to Gaussian processes. 
	The main contribution is the construction of variational ensembles given a dictionary of fitted Gaussian processes without revisiting any subset of observations.
	Our framework allows for regression, classification and
        heterogeneous tasks, i.e.\ mix of continuous and
          discrete variables over the same input domain.
	We exploit infinite-dimensional integral operators based on the Kullback-Leibler divergence between stochastic processes to re-combine arbitrary amounts of variational sparse approximations with different complexity, likelihood model and location of the pseudo-inputs. 
	Extensive results illustrate the usability of our framework in
        large-scale distributed experiments,
        also compared with the exact inference models in the literature.
\end{abstract}

\section{Introduction}

\begin{wrapfigure}[17]{r}{0.33\textwidth}
	\centering
	\vspace{-0.5cm}%
	\hspace{-4mm}%
	\includegraphics[width=0.31\textwidth]{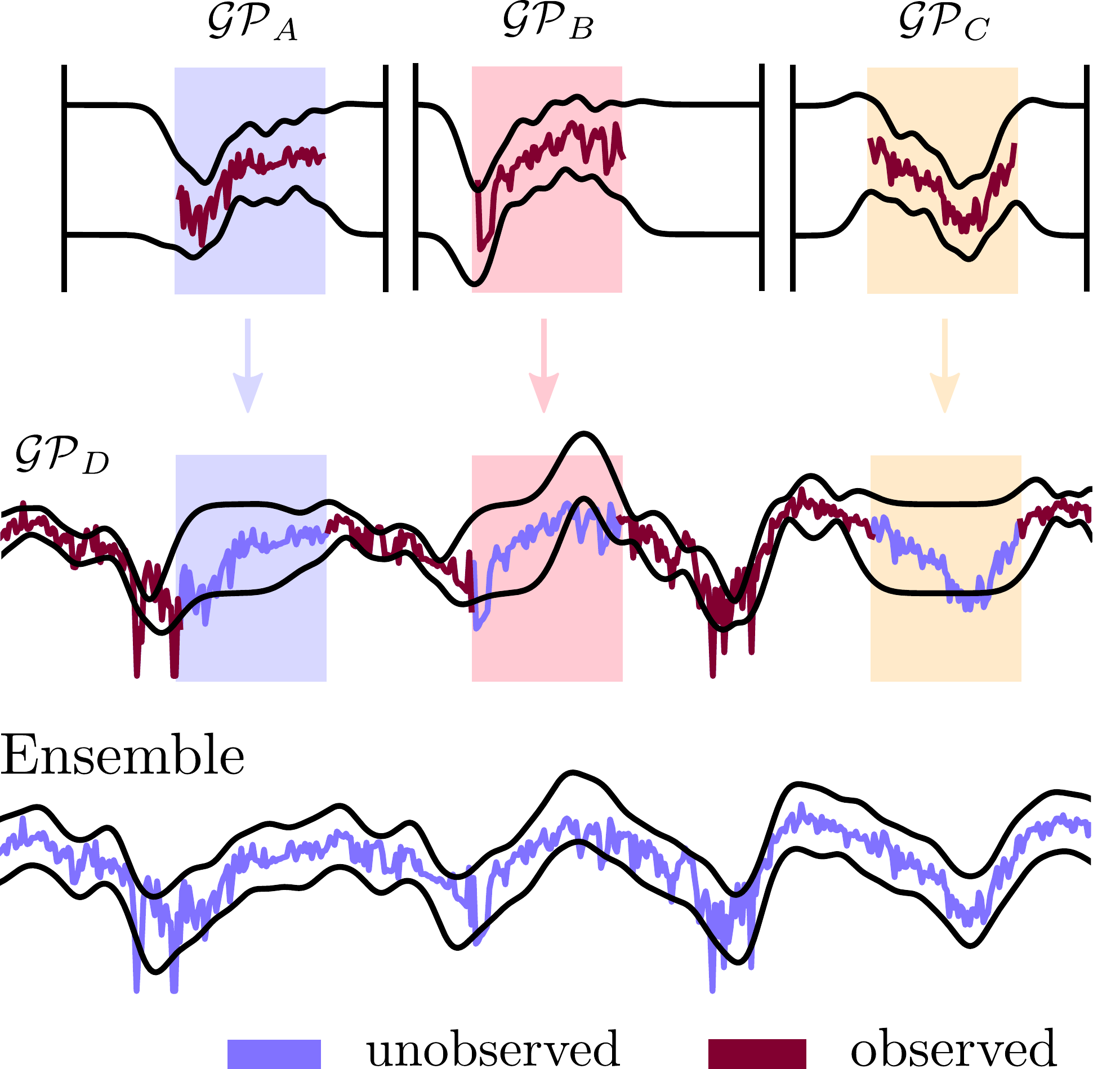} 
	\caption{Recyclable GPs (\emph{A}, \emph{B}, \emph{C} and \emph{D}) are re-combined without accessing to the subsets of observations.}
	\label{fig:graphic}
\end{wrapfigure}
% Paragraph 1.
One of the most desirable properties for any modern machine learning method is the handling of very large datasets. 
Since this goal has been progressively achieved in the literature with scalable models, much attention is now paid to the notion of efficiency.
For instance, in the way of accessing data.
The fundamental assumption used to be that samples can be revisited without restrictions \textit{a priori}. 
In practice, we encounter cases where the massive storage or data centralisation is not possible anymore for preserving the privacy of individuals, e.g. health and behavioral data.
%
% must be avoided for preserving the privacy of individuals.
%
The mere limitation of data availability forces learning algorithms to derive new capabilities, such as i) distributing the data for \textit{federated learning} \citep{smith2017federated}, ii) observe streaming samples for \textit{continual learning} \citep{goodfellow2014empirical} and iii) limiting data exchange for \mbox{\textit{private-owned models} \citep{peterson2019private}.} %\citep{dwork2014algorithmic}. 

A common theme in the previous approaches is
the idea of model memorising and recycling, i.e.\ using the already fitted parameters in another problem or joining it with others for an additional global task without revisiting any data. 
If we look to the functional view of this idea, uncertainty is still much harder to be repurposed than parameters. This is the point where Gaussian process (GP) models \citep{williams2006gaussian} play their role.
%
%Significant advances have been achieved in the functional view of this idea, typically with neural networks (NNs) based on regularisation for assuring that the underlying function learning task is preserved. 
%However, the function uncertainty is still much harder to be propagated forward than parameters, here is where Gaussian process (GP) models \citep{williams2006gaussian} play their role.

In this paper, we investigate a general framework for recycling
distributed variational sparse approximations to GPs, illustrated in
Figure \ref{fig:graphic}. Based on the properties of the Kullback-Leibler divergence between stochastic processes
\citep{matthews2016sparse} and Bayesian inference, our method ensembles an arbitrary
amount of variational GP models with different complexity, likelihood and location of pseudo-inputs, without revisiting any data.

%\subsection{Background}

\textbf{1.1~~~Background.~} The flexible nature of GP models for definining prior distributions over non-linear function spaces has made them a suitable alternative in many probabilistic regression and classification problems. However, GP models are not immune to settings where the model needs to adapt to \textit{irregular} ways of accessing the data, e.g. asynchronous observations or missings input areas. Such settings, together with GP model's well-known
computational cost for the exact solutions, typically $\Ocal(N^3)$
where $N$ is the data size, has motivated plenty of
aproaches focused on \emph{parallelising inference}.
%
%Their well-known computational cost for exact inference, typically $\Ocal(N^3)$ where $N$ is the data size, has motivated plenty of GP alternatives. Apart from the low rank approximations \citet{quinonero2005unifying,snelson2006sparse} based on $M$ inducing variables, where $M\ll N$, or via the Nystr\"om method \citep{williams2001using} for speeding up the inversion of covariance matrices, there has been relevant works focused on the parallelisation of calculus for inference purposes. 
%
Regarding the task of distributing the computational load between
learning agents, GP models have been inspired by \emph{local
    experts} \citep{jacobs1991adaptive,hinton2002training}. %This link between neural network (NN) experts and GPs is understandable as one infinitely wide, the NN, can be seen as the other \citep{neal1996bayesian}. 
Two seminal works exploited this connection before the modern
era of GP approximations. While the Bayesian committee machine (BCM)
of \citet{tresp2000bayesian} focused on merging  independently trained
GP regression models on subsets of the same data, the infinite mixture
of GP experts \citep{rasmussen2002infinite} increased the model expresiveness by combining local GP experts. Our proposed method will be closer to the first approach whilst the second one is also amenable but out of the spirit of this work. 

The emergence of large datasets, with size $N$$>$$10^{4}$, led to the introduction of approximate models, that in combination with variational inference \citep{titsias2009variational}, succeed in scaling up GPs. % has been highly influential in the goal of scaling up GPs. 
Two more recent
approaches that combine sparse GPs with ideas from distributed models
or computations are \citet{gal2014distributed} and
\citet{deisenroth2015distributed}.  Based on the variational GP
inference of \citet{titsias2009variational},
\citet{gal2014distributed} presented a new re-parameterisation of the
lower bounds that allows to distribute the computational load accross
nodes, also applicable to GPs with stochastic variational inference
\citep{hensman2013gaussian} and with non-Gaussian likelihoods
\citep{hensman2015scalable,saul2016chained}. Out of the sparse GP approach and more inspired in
\citet{tresp2000bayesian} and product of experts
\citep{bordley1982multiplicative}, the distributed GPs of
\citet{deisenroth2015distributed} scaled up the parallelisation
mechanism of local experts to the range of $N$$>$$10^6$. Their
approach is focused on exact GP regression, not considering
classification or other non-Gaussian likelihoods. Table
\ref{tab:models} provides a description of these different methods and
their main properties, also if each distributed node is a GP model itself.
%We are interested in the case of having multiple subsets of data that are used in parallel for training independent variational GP models, one per subset, being either for a regression or classification task. The key idea that we explore is that, if assumming some stationary conditions and having a dictionary of \textit{already} learned variational local experts, it would be possible to obtain a variational global expert by combining them but without re-visiting any data.
%
\begin{table}[ht!]
	\vspace*{-\baselineskip}
	\caption{Properties of distributed GP models}
	\centering
	\resizebox{\textwidth}{!}{%
	\begin{tabular}{cccccccc}
		\toprule
		\textsc{Model} & $\mathcal{N}$ \textsc{reg.} & non-$\mathcal{N}$ \textsc{reg.} &\textsc{class.} & \textsc{het.} & \textsc{inference}& $\mathcal{GP}_{\textsc{node}}$ & \textsc{data st.}\\
		\midrule
		\citet{tresp2000bayesian}& \cmark & \xmark & \xmark & \xmark &Analytical & \cmark & \cmark\\
		\citet{ng2014hierarchical} & \cmark & \xmark & \xmark & \xmark &Analytical & \cmark & \cmark\\
		\citet{cao2014generalized} & \cmark & \xmark & \xmark & \xmark &Analytical & \cmark & \cmark\\
		\citet{deisenroth2015distributed} & \cmark & \xmark & \xmark & \xmark &Analytical & \cmark & \cmark\\
		\citet{gal2014distributed} & \cmark & \cmark & \cmark &  \xmark&Variational & \xmark & \xmark\\
		\textbf{This work} & \cmark &\cmark & \cmark & \cmark &Variational & \cmark & \xmark\\
		\bottomrule
	\end{tabular}}
{\raggedright \scriptsize $(^{*})$ Respectively, Gaussian and non-Gaussian regression ($\mathcal{N}$ \& non-$\mathcal{N}$ \textsc{reg}), classification (\textsc{class}), heterogeneous (\textsc{het}) and storage (\textsc{st}).\par}
%{\raggedright \scriptsize ~~~~~~ heterogeneous (\textsc{het}) and storage (\textsc{st}).\par}
\label{tab:models}
\vspace*{-\baselineskip}
\end{table}

Our contribution in this paper is to provide a new framework for recycling already fitted variational approximations with the purpose of building a global GP ensemble. We use infinite-dimensional integral operators, that can be understood in the context of the Kullback-Leibler (KL) divergence between stochastic processes \citep{matthews2016sparse}. 
%
%This idea was also considered in \citep{bui2017streaming,moreno2019continual} for the adaptation of GPs to continual learning. 
%
Here, we borrow the reparameterisation of \citet{gal2014distributed}
to distribute the inference computation. The recyclable framework is
amenable for both regression, classification and heterogeneous tasks,
e.g.\ outputs are a mix of continuous and discrete variables. We are
not restricted to any specific sparse GP approach, and several of them
are suitable for combination under the ensemble GP, for instance,
\citet{hensman2013gaussian,hensman2015scalable,saul2016chained}. We
experimentally provide evidence of the performance of the recyclable
GP framework in different settings and under several architectures. Our model can be viewed as an extension of
\citet{gal2014distributed} and \citet{deisenroth2015distributed} but
generalised to multiple variational models without restrictions on the
learning task or the likelihood distribution adopted. %We provide a Pytorch code that allows to easily building ensembles of GP models.\footnote{The code is publicly available in the repository: \texttt{github.com/xxxxxxxx/RecyclableGP/}.} The syntax follows the idea of providing a list~ \texttt{recyclable\_models = [$\texttt{GP}_1$, $\texttt{GP}_2$, $\texttt{GP}_3$]} for the ensemble, where each $\texttt{GP}_k$ contains uniquely the variational parameters and hyperparameters learned by the independent local approximations.
\vspace*{-0.5\baselineskip}
\section{Recyclable Gaussian Processes}
%\vspace*{-0.5\baselineskip}
We consider a supervised learning problem, where we have an
input-output training dataset $\Dcal = \{\xc_{i}, y_i\}^{N}_{i=1}$
with $\xc \in \mathbb{R}^{p}$. We assume i.i.d.\ outputs $y_i$, that can be either continuous or discrete variables. %Typically, in the exact GP regression case, the likelihood distribution is $\mathcal{N}(y|\mu, \sigma^2_n)$, where $\mu = f(\xc)$, being $f(\cdot)$ a non-linear function. Independently of the likelihood model, $f$ is assumed to be drawn from a zero-mean GP prior $f \sim \mathcal{GP}(\bm{0}, k(\cdot, \cdot))$. 
For convenience, we will refer to the likelihood term
$p(y|\bm{\theta})$ as $p(y|f)$ where the generative parameters are
linked via $\bm{\theta} = f(\xc)$, being $f(\cdot)$ a non-linear
function drawn from a zero-mean GP prior $f \sim
  \mathcal{GP}(\bm{0}, k(\cdot, \cdot))$, and $k(\cdot, \cdot)$ is the covariance function or \textit{kernel}.  Importantly, when non-Gaussian outputs are considered, the GP output function $f(\cdot)$ might need an extra deterministic mapping $\Phi(\cdot)$ that transforms it to the appropriate parametric domain of $\bm{\theta}$. 

The data $\Dcal$ is assumed to be partitioned into an arbitrary number
of $K$ subsets that we aim to observe and process independently, that
is, $\{\Dcal_1, \Dcal_2, \dots, \Dcal_K\}$. There is not any
restriction on the amount of subsets or learning
nodes. The subsets $\{\Dcal_k\}_{k=1}^K$ do not need to have the same size, and we only restrict them to be $N_k$$<$$N$. However, since we treat with a huge number of observations, we still consider that $N_k$ for all $k\in\{1,2,\dots,K\}$ is sufficiently large for not accepting exact GP inference due to temporal and computational demand. Notice that  $k$ is an index while $k(\cdot,\cdot)$ refers to the \textit{kernel}.
\subsection{Sparse variational approximations for distributed subsets}
\vspace*{-0.25\baselineskip}
We adopt the sparse GP approach based on \emph{inducing variables},
together with the variational framework of
  \citet{titsias2009variational} for the partitioned subsets. The use of auxiliary variables within approximate
inference methods is widely known in the GP literature
\citep{hensman2013gaussian,hensman2015scalable,matthews2016sparse}. %,burt2019rates}.
In the context of $K$ distributed partitions and their adjacent samples, we define subsets of $M_k$$\ll$$N_k$ inducing inputs $\bm{Z}_k = \{\zc_{m}\}^{M_k}_{m=1}$, where $\zc_{m} \in \mathbb{R}^p$ and their non-linear function evaluations by $f(\cdot)$ are denoted as $\uc_k = [f(\zc_{1}), f(\zc_{2}), \cdots, f(\zc_{M_k})]^{\top}$. We remark that $f$ is considered to be stationary across all distributed tasks, being $\uc_k$ $\forall k \in \{1,2,\dots, K\}$ values of the same function.

To obtain multiple independent approximations to the posterior distribution $p(f|\Dcal)$ of the GP function, we introduce $K$ variational distributions $q_k(f)$, one per distributed partition $\Dcal_k$. In particular, each variational distribution factorises as $q_k(f) = p(f_{\neq \uc_k}| \uc_k)q_k(\uc_k)$, with $q_k(\uc_k) = \mathcal{N}(\uc_k|\bm{\mu}_k, \bm{S}_k)$ and $p(f_{\neq \uc_k}| \uc_k)$ being the standard conditional GP prior distribution given the hyperparameters $\bm{\psi}_k$ of each $k$-th kernel. %Moreover, as variational approximation might lead to intractable computations due to non-Gaussian likelihood models.
To fit the local variational distributions $q_k(\uc_k)$, we build lower bounds $\Lcal_k$ on the marginal log-likelihood (ELBO) of every data partition $\Dcal_k$.
Then, we use optimisation methods, typically gradient-based, to maximise the $K$ objective functions $\Lcal_k$, one per distributed task, separately. Each local ELBO is obtained as follows
\begin{equation}
\label{eq:local}
	\Lcal_k = \sum_{i=1}^{N_k}\mathbb{E}_{q_k(\f_i)}\left[\log p(y_i|\f_i)\right] - \text{KL}[q_k(\uc_k)||p_k(\uc_k)],
\end{equation}
with $p_k(\uc_k) = \mathcal{N}(\uc_k|\bm{0}, \K_{kk})$, where $\K_{kk} \in \mathbb{R}^{M_k\times M_k}$ has entries $k(\zc_{m}, \zc_{m'})$ with $\zc_{m}, \zc_{m'} \in \bm{Z}_k$ and conditioned to certain kernel hyperparameters $\bm{\psi}_k$ that we also aim to optimise. The variable $\f_i$ corresponds to $f(\xc_i)$ and the  marginal posterior comes from $q_k(\f_i) = \int p(\f_i|\uc_k)q_k(\uc_k)d\uc_k$. In practice, the distributed local bounds $\Lcal_k$ are identical to the one presented in \citet{hensman2015scalable} and also accept stochastic variational inference \citep{hoffman2013stochastic,hensman2013gaussian}. An important detail is that, while the GP function is restricted to be stationary between tasks, the likelihood distribution model $p(y_i|f)$ is not. An example of the heterogeneous setting is shown the experiments, where we combine a Gaussian and a Bernoulli likelihood.

 %based on smaller \emph{mini-batches} of partitions. Notice that this process let the distributed nodes to maximise the target functions much faster and indeed, to obtain fitted solutions to $q_k(\uc_k)$ in an asynchronous way. An important detail is that, while the GP function is restricted to be stationary between tasks, the likelihood distribution model $p(y_i|f)$ is not. We show an example of the heterogeneous setting in the experiments, where we combine a Gaussian and a Bernoulli likelihood.
\vspace*{-0.25\baselineskip}
\subsection{Global inference from local learning}
\vspace*{-0.25\baselineskip}
Having a \emph{dictionary} which contains the already fitted local
variational solutions, while others can be still under computation, we focus on how using them for performing global inference of the GP. Such dictionary consists, for instance, of a list of objects $\Ecal = \{\Ecal_1, \Ecal_2, \dots, \Ecal_K\}$ without any specific order, where each $\Ecal_k = \{\bm{\phi_k}, \bm{\psi_k}, \bm{Z}_k\}$, $\bm{\phi_k}$ being the corresponding variational parameters $\bm{\mu}_k$ and $\bm{S}_k$.

Ideally, to obtain a global inference solution given the GP models
included in the dictionary, the resulting posterior distribution
should be valid for all the local subsets of data. This is only
possible if we consider the entire data set $\Dcal$ in a maximum
likelihood criterion setting. Specifically, our goal now is to obtain
an approximate posterior $q(f)\approx p(f|\Dcal)$ by maximising a
lower bound $\Lcal_{\Ecal}$ under the log-marginal likelihood $\log
p(\Dcal)$ without revisiting the data already observed by the local
models. We begin by considering the full posterior distribution of the
stochastic process, similarly as \citet{burt2019rates} does for
obtaining an upper bound on the KL divergence. The idea is to use infinite-dimensional integral operators that were introduced by
\citet{matthews2016sparse} in the context of variational inference,
and previously by \citet{seeger2002pac} for standard GP error
bounds. The use of the infinite-dimensional integrals is equivalent to
an \emph{augment-and-reduce} strategy \citep{ruiz2018augment}. It consists of two steps: i) we augment the model to accept the conditioning on the infinite-dimensional stochastic process and ii) we use properties of Gaussian marginals to reduce the infinite-dimensional integral operators to a finite amount of GP function values of interest. Similar strategies have been used in continual learning for GPs \citep{bui2017streaming,moreno2019continual}. 

\textbf{Global objective.}~~The construction considered is as follows. We first denote $\yc$ as all the output targets $\{y_i\}^N_{i=1}$in the dataset $\Dcal$ and $f_{\infty}$ as the \emph{augmented} infinite-dimensional GP. Notice that $f_{\infty}$ contains all the function values taken by $f(\cdot)$, including that ones at $\{\xc_i\}^{N}_{i=1}$ and $\{\bm{Z}_k\}^{K}_{k=1}$ for all partitions. The augmented log-marginal expression is therefore 
\begin{equation}
\label{eq:marginal}
	 \log p(\yc) = \log p(\yc_1, \yc_2, \dots, \yc_K) = \log \int p(\yc, f_{\infty})df_{\infty}, 
\end{equation}
where each $\yc_k = \{y_i\}^{N_k}_{i=1}$ is the subset of output
values already used for training the local GP models. The joint
distribution in \eqref{eq:marginal} factorises according to $p(\yc,
f_{\infty}) = p(\yc| f_{\infty})p( f_{\infty})$, where the l.h.s.\
term is the augmented likelihood distribution and the r.h.s.\ term
would correspond to the full GP prior over the entire stochastic
process. Then, we introduce a global variational distribution
$q(\bm{u_*}) = \mathcal{N}(\bm{u_*}|\bm{\mu_*}, \bm{S_*})$ that we aim
to fit by maximising a lower bound under $\log p(\yc)$. The variables
$\up$ correspond to function values of
$f(\cdot)$ given a new subset of inducing inputs $\bm{Z_{*}} =
\{\zc_m\}^{M}_{m=1}$, where $M$ is the free-complexity degree of the
global variational distribution.  To derive the bound, we exploit the
reparameterisation introduced by \citet{gal2014distributed} for
distributing the computational load of the expectation term. It is
based in a double application of the Jensen's inequality and obtained as
\begin{multline}
\label{eq:bound1}
\log p(\yc) = \log \iint q(\up)p(f_{\infty\neq\up}|\up)p(\yc|\finf)\frac{p(\up)}{q(\up)}df_{\infty\neq\up}d\up\\
\geq \Ebb_{q(\up)} \left[\Ebb_{p(\fc_{\infty\neq \up}|\up)}\left[ \log p(\yc | \finf)\right] +\log \frac{p(\up)}{q(\up)}  \right],
\end{multline}
where we applied the properties of Gaussian conditionals to factorise the GP prior as $p(\finf) = p(\fc_{\infty\neq \up}|\up)p(\up)$. Here, the last prior distribution is $p(\up) = \mathcal{N}(\up|\bm{0}, \Kpp)$ where $\left[\Kpp\right]_{m,n} := k(\zc_m, \zc_n)$, with $\zc_m, \zc_n \in \bm{Z_{*}}$, conditioned to the global kernel hyperparameters $\bm{\psi_*}$ that we also aim to estimate. The double expectation in \eqref{eq:bound1} comes from the factorization of the infinite-dimensional integral operator and the application of the Jensen's inequality twice. Its derivation is in the appendix.

\textbf{Local likelihood reconstruction.}~ The augmented likelihood distribution is perhaps, the most important point of the derivation. It allows us to apply conditional independence (CI) between the subsets of distributed output targets. This gives a factorized term that we will later use for introducing the local variational experts in the bound, that is, $\log p(\yc | \finf) = \sum_{k=1}^{K}\log p(\yc_k| \finf)$. To avoid revisiting local likelihood terms, and hence, evaluating distributed subsets of data that might not be available, we use the Bayes theorem but conditioned to the infinite-dimensional augmentation. It indicates that the local variational distributions can be approximated as
\begin{equation}
\label{eq:bayes}
	q_k(\finf) \approx p(\finf|\yc_k) \propto p(\finf)p(\yc_k|\finf),
\end{equation}
where the augmented approximate distribution factorises as
$q_k(\finf) = p(f_{\infty\neq\uk}|\uk)q_k(\uk)$ as in the
variational framework of \citet{titsias2009variational}. Similar
expressions consisting on the full stochastic process conditionals
were previously used in \citet{bui2017streaming} and
\citet{matthews2016sparse}, with emphasis on the theoretical
  consistency of augmentation. Thus, we can find an approximation for each local likelihood term $p(\yc_k|\finf)$ by inverting the Bayes theorem in \eqref{eq:bayes}. Then, the conditional expectation in \eqref{eq:bound1} turns to be
\begin{equation*}
\Ebb_{p(\fc_{\infty\neq \up}|\up)}\left[\log p(\yc| \finf)\right] \approx \sum_{k=1}^{K} \Ebb_{p(\fc_{\infty\neq \up}|\up)}\left[\log \frac{q_k(\finf)}{p(\finf)}\right]
= \sum_{k=1}^{K} \Ebb_{p(\uk|\up)}\left[ \log\frac{q_k(\uk)}{p(\uk)}\right],
\end{equation*}
%\vspace*{-0.25\baselineskip}
where we applied properties of Gaussian marginals to \textit{reduce}
the infinite-dimensional expectation, and factorised the distributions to be explicit on each subset of fixed inducing-variables $\uk$ rather than $\finf$. For instance, the integral $\int p(\finf) d\fc_{\infty\neq \uk}$ is analogous to $\int p(\fc_{\infty\neq \uk}, \uk) d\fc_{\infty\neq \uk} = p(\uk)$ via marginalisation.

\textbf{Variational contrastive expectations.}~ The introduction of $K$ expectation terms over the log-ratios in the bound of \eqref{eq:bound1} as a substitution of the local likelihoods, leads to particular advantages. If we have a \textit{nested} integration in \eqref{eq:bound1}, first over $\up$ at the conditional prior distribution, and second over $\uk$ given the log-ratio $q_k(\uk)/p(\uk)$, we can exploit the GP predictive equation to write down
%\vspace*{-0.25\baselineskip}
\begin{equation}
\label{eq:contrastive}
\sum_{k=1}^{K} \Ebb_{q(\up)} \left[\Ebb_{p(\uk|\up)}\left[ \log \frac{q_k(\uk)}{p(\uk)}\right]\right] = \sum_{k=1}^{K} \Ebb_{\qt}\left[ \log \frac{q_k(\uk)}{p(\uk)} \right],
\end{equation}
where we obtained $\qt$ via the integral $\qt = \int
q(\up)p(\uk|\up)d\up$, that coincides with the approximate
predictive GP posterior. This
distribution can be obtained analytically for each $k$-th subset $\uk$
using the following expression, whose complete derivation is provided in the appendix,
\begin{equation*}
\qt = \Ncal(\uk|\Kpk^{\top}\Kpp^{-1}\mup, \K_{kk} + \Kpk^{\top}\Kpp^{-1}(\Sp - \Kpp)\Kpp^{-1}\Kpk),
\end{equation*}
where, once again, $\bm{\phi_*} =\{\mup, \Sp\}$ are the global variational parameters that we aim to learn.
One important detail of the sum of expectations in \eqref{eq:contrastive} is that it works as an average contrastive indicator that measures how well the global $q(\up)$ is being fitted to the local experts $q_k(\uk)$. Without the need of revisiting any distributed subset of data samples, the GP predictive $\qt$ is playing a different role in contrast with the usual one. Typically, we assume the approximate posterior fixed and fitted, and we evaluate its performance on some test data points. In this case, it goes in the opposite way, the approximate variational distribution is unfixed, and it is instead evaluated over each $k$-th local subset of inducing-inputs $\bm{Z}_k$.

%\subsection{Lower ensemble bounds}

\textbf{Lower ensemble bounds.}~ We are now able to simplify the initial bound in \eqref{eq:bound1} by substituting the first term with the contrastive expectations presented in \eqref{eq:contrastive}. This substitution gives the final version of the lower bound $\Lcal_{\Ecal}\leq \log p(\yc)$ on the log-marginal likelihood for the global GP,
\begin{equation}
	\label{eq:ensemble}
	\Lcal_{\Ecal} = \sum_{k=1}^{K} \Ebb_{\qt}\left[ \log q_k(\uk) - \log p(\uk) \right] - \text{KL}\left[q(\bm{u_*})|| p(\bm{u_*})\right].
\end{equation}
The maximisation of \eqref{eq:ensemble} is w.r.t. the
parameters $\bm{\phi_*}$, the hyperparameters $\bm{\psi_*}$ and
$\bm{Z_*}$. To assure the positive-definitiness of variational
covariance matrices $\{\bm{S}_k\}^K_{k=1}$ and $\bm{S_*}$ on both
local and global cases, we consider that they all
factorize according to the Cholesky decomposition $\bm{S} =
\L\L^{\top}$. We can then use unconstrained optimization to find
optimal values for the lower-triangular matrices $\L$. 

A priori, the ensemble GP bound is agnostic with respect to the
likelihood model. There is a general derivation in
\citet{matthews2016sparse} of how stochastic processes and their integral operators are affected by projection
functions, that is, different linking mappings of the function
$f(\cdot)$ to the parameters $\bm{\theta}$. In such cases,
the local lower bounds $\Lcal_k$ in \eqref{eq:local} might
include expectation terms that are intractable. Since we build the framework to accept any possible data-type, we propose to solve the integrals via Gaussian-Hermite quadratures as in \citet{hensman2015scalable,saul2016chained} and if this is not possible, an alternative would be to apply Monte-Carlo methods.

\textbf{Computational cost and connections.~} The computational cost of the local models is $\Ocal(N_kM_k^{2})$, while the global GP reduces to $\Ocal((\sum_kM_k)M^{2})$ and  $\Ocal(M^{2})$ in training and prediction, respectively. The methods in Table \ref{tab:models} typically need $\Ocal(\sum_k N_k^{2})$ for global prediction. A last theoretical aspect is the link between the global bound in \eqref{eq:ensemble} and the underlying idea in \citet{tresp2000bayesian,deisenroth2015distributed}. Distributed GP models are based on the application of CI to factorise the likelihood terms of subsets. To approximate the posterior predictive, they combine local estimates, divided by the GP prior. It is analogous to \eqref{eq:ensemble}, but in the logarithmic plane and the variational inference setup.
\vspace*{-0.5\baselineskip}
\subsection{Capabilities of recyclable GPs}

We highlight several use cases for the proposed framework. The idea of recycling GP models opens the door to multiple extensions, with particular attention to the local-global modelling of heterogeneous data problems and the adaptation of model complexity in a data-driven manner.%, e.g. dimension of $M$.

\textbf{Global prediction.~} Our purpose might be to predict how
likely an output test datum $y_t$ is at some point $\xc_t$ of the
input space $\mathbb{R}^{p}$. In this case, the global predictive
distribution can be approximated as $p(y_t|\Dcal) \approx \int
p(y_t|\f_t)q(\f_t)d\f_t$, with $ q(\f_t) = \int
p(\f_t|\up)q(\up)d\up$. As mentioned in the previous subsection, the integral can be obtained by quadratures  when the solution is intractable.

\textbf{Heterogeneous single-output GP.~} Extensions to GPs with
heterogeneous likelihoods, that is, a mix of continuous and
discrete variables $\yc_i$, have been proposed for
multi-output GPs \citep{moreno2018heterogeneous}. However, there are
no restrictions in our single-output model to accept
different likelihoods $p(y_i|f(\xc_i))$ per data point $\{\xc_i,y_i\}$. An inconvenience of the bound in \eqref{eq:local}, is that, each $i$-th expectation term could be imbalanced with respect to the others. For example, if mixing Bernoulli and Gaussian variables,
binary outputs could contribute more to the objective function than the rest, due to the dimensionality. To overcome this
issue, we fit a local GP model to each heterogeneous variable. We join all models together using the ensemble
bound in \eqref{eq:ensemble} to propagate the uncertainty in a principled way. Although, data-types need to be known beforehand, perhaps as additional \textit{labels}. %our framework will propagate the uncertainty of the local GP models in a principled way.

\textbf{Recyclable GPs and new data.~} In practice, it might not be 
necessary to distribute the whole dataset $\Dcal$ in parallel tasks, with
some subsets $\Dcal_k$ available at the global ensemble. It is possible to
combine the samples in $\Dcal_k$ with the dictionary of local GP
variational distributions. In such cases, we would only approximate the
likelihood terms in \eqref{eq:bound1} related to the distributed
subsets of samples. The resulting \textit{combined} bound would be
equivalent to \eqref{eq:ensemble} with an additional expectation term
on the new data. We provide the derivation of this \textit{combined}
bound in the supplementary material.
%\textbf{Stationary conditions.}~~ Stationary kernels. If non-stationary, use forgetting strategies.

\textbf{Stationarity and expressiveness.~} We assume that the
non-linear function $f$ is stationary across subsets of data. If this
assumption is relaxed, some form of adaptation or \textit{forgetting} should be included to
match the local GPs. Other types of models can be considered
for the ensemble, as for instance, with several latent functions
\citep{lazaro2011variational} or sparse multi-output
GPs \citep{alvarez2011computationally}. The
model also accepts GPs with increased expressiveness. For example, to get
multi-modal likelihoods, we can use mixture of GP
experts \citep{rasmussen2002infinite}.

\textbf{Data-driven complexity and recyclable ensembles.~} One of the main advantages of the recyclable GP framework is that it allows \textit{data-driven} updates of the complexity. That is, if an ensemble ends in a new variational GP model, it also can be recycled. Hence, the number of global inducing-variables $M$ can be iteratively increased conditioned to the amount of samples considered. A similar idea was already commented as an application of the sparse order-selection theorems by \citet{burt2019rates}. 

\textbf{Model recycling and use cases.~} The ability of recycling GPs in future global tasks  have a significant impact in behavioral applications, where fitted private-owned models in smartphones can be shared for global predictions rather than data. Its application to medicine is also of high interest. If one has personalized GPs for patients, epidemiologic surveys can be built without centralising private data.
\vspace*{-0.75\baselineskip}
\section{Related Work}
\vspace*{-0.5\baselineskip}
In terms of distributed inference for scaling up computation, that is,
the delivery of calculus operations  across parallel nodes but not
data or independent models, we are similar to
\citet{gal2014distributed}. Their approach can be understood as a
specific case of our framework. Alternatively, if we look to the
property of having nodes that contain usable GP models (Table
\ref{tab:models}), we are similar to
\citet{deisenroth2015distributed,cao2014generalized} and
\citet{tresp2000bayesian}, with the difference that we introduce
variational approximation methods for non-Gaussian likelihoods. An
important detail is that the idea of exploiting properties of full
stochastic processes \citep{matthews2016sparse} for substituting
likelihood terms in a general bound has been previously considered in
\citet{bui2017streaming} and \citet{moreno2019continual}. Whilst the
work of \citet{bui2017streaming} ends in the derivation of
expectation-propagation (EP) methods for streaming inference in GPs,
the introduction of the reparameterisation of
\citet{gal2014distributed} makes our inference and performance
different from \citet{moreno2019continual}. There is also the inference framework of \citet{bui2018partitioned} for both federated and continual learning, but focused on EP and the Bayesian approach of \citet{nguyen2017variational}. A short analysis of its application to GPs is included for continual learning settings but far from the large-scale scope of our paper. 
Moreover, the spirit of using inducing-points as pseudo-approximations
of local subsets of data is shared with \citet{bui2014tree}, that
comments its potential application to distributed setups. More oriented to dynamical modular models, we find the work by
\citet{velychko2018making}, whose factorisation across tasks is
similar to \citet{ng2014hierarchical} but oriented to state-space models. 
\vspace*{-0.75\baselineskip}
\section{Experiments}
\vspace*{-0.5\baselineskip}
In this section, we evaluate the performance of our framework for multiple recyclable GP models and data access settings. To illustrate its usability, we present results in three different learning scenarios: i) regression, ii) classification and iii) heterogeneous data. All experiments are numbered from one to nine in roman characters. Performance metrics are given in terms of the negative log-predictive density (\textsc{nlpd}), root mean square error (\textsc{rmse}) and mean-absolute error (\textsc{mae}). We provide Pytorch code that allows to easily learn the GP ensembles.\footnote{The code is publicly available in the repository: \texttt{github.com/pmorenoz/RecyclableGP/}.} It also includes the baseline methods. The syntax follows the spirit of providing a list of~ \texttt{recyclable\_models = [$\texttt{GP}_1$, $\texttt{GP}_2$, $\texttt{GP}_3$]}, where each $\texttt{GP}_k$ contains exclusively parameters of the local approximations. Further details about initialization, optimization and metrics are in the appendix. Importantly, we remark that data is \underline{never revisited} and its presence in the ensemble plots is just for clarity in the comprehension of results.

%Remark that data is NOT re-visited and neither used again. Only plotted for clarification.

\begin{figure}[ht]
	\begin{minipage}[b]{0.5\linewidth}
		\centering
		\caption{Recyclable GPs with synthetic data.}
		\includegraphics[width=1\linewidth]{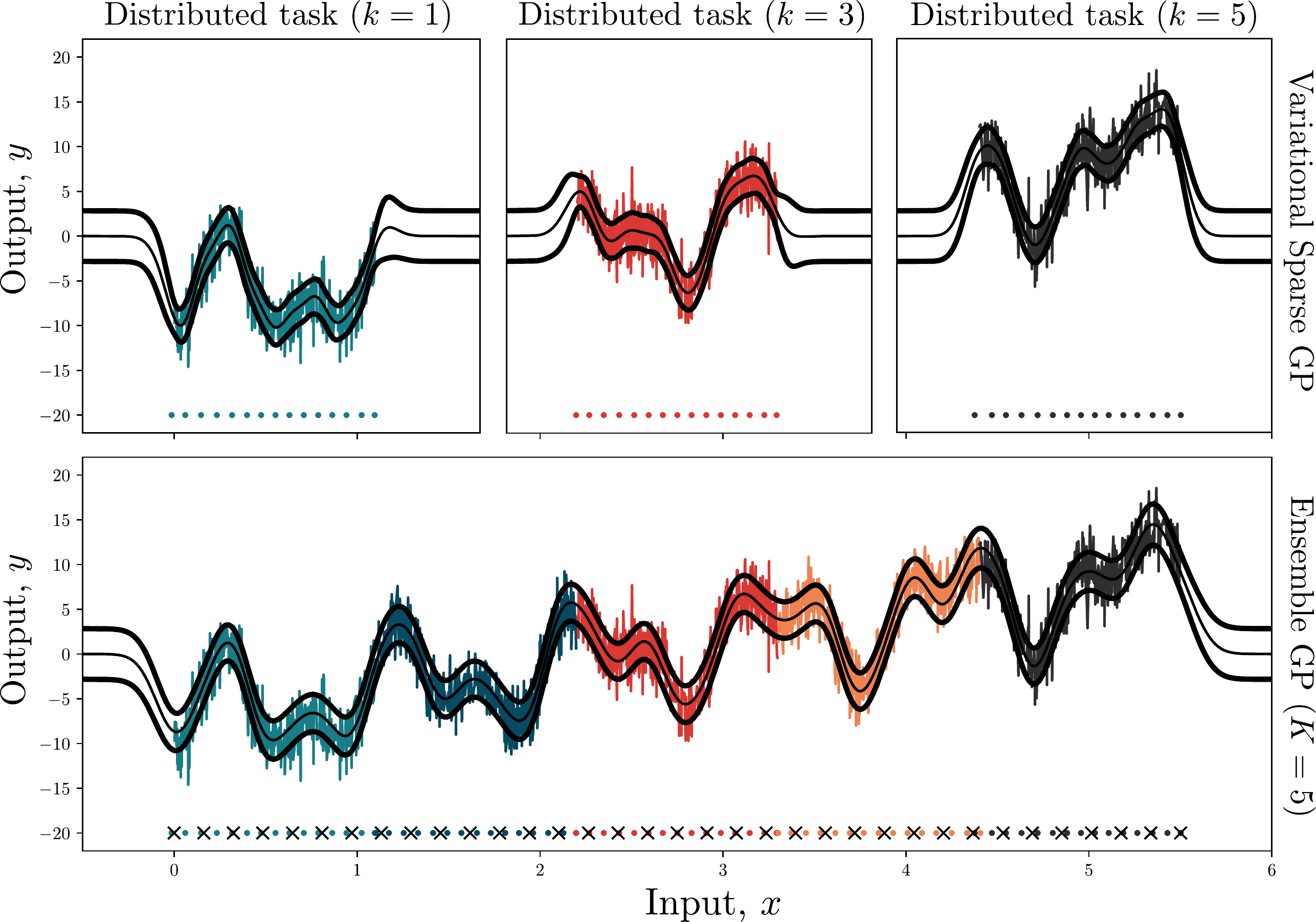}
		\label{fig:parallel}
	\end{minipage}
\hspace*{0.5cm}
\begin{minipage}[b]{0.35\linewidth}
	\centering
	\renewcommand{\figurename}{Table}
	\renewcommand{\thefigure}{2}
	\caption{Performance metrics for distributed GP regression with the solar physics dataset. (std. $\times 10^{2}$)}
	\renewcommand{\figurename}{Figure}
	\renewcommand{\thefigure}{\arabic{figure}}
	\resizebox{\textwidth}{!}{
		\begin{tabular}{ccc}
		\toprule
		\textsc{Model}&\textsc{nlpd}&\textsc{rmse}\\
		\midrule
		BCM&$-$&$17.25$\\
		PoE&$1.51\pm0.01$&$1.08$\\
		GPoE&$1.51\pm0.07$&$1.08$\\
		RBCM&$1.53\pm0.01$&$1.11$\\
		%DVIGP&$\pm$&$\pm$\\
		\textbf{This work}&$1.68\pm0.14$&$1.17\pm0.12$\\
		\bottomrule
	\end{tabular}}
\vspace*{2.3cm}
\label{tab:solar}
\end{minipage}
%\vspace*{-\baselineskip}
\end{figure}

\textbf{4.1~~~Regression.~} In our first experiments for variational
GP regression with distributed models, we provide both qualitative and
quantitative results about the performance of recyclable
ensembles. 
\textbf{(i) Toy concatenation:} In Figure \ref{fig:parallel}, we
show three of five tasks united in a new GP model.
Tasks are GPs fitted independently with $N_k$$=$$500$ synthetic data
points and $M_k$$=$$15$ inducing variables per distributed task. The
ensemble fits a global variational solution of dimension
$M$$=$$35$. Notice that the global variational GP tends to match the uncertainty of the local approximations. 
\textbf{(ii) Distributed GPs:} We provide error
metrics for the recyclable GP framework compared with the
state-of-the-art models in Table \ref{tab:metrics}. The training data
is synthetic and generated as a combination of $\sin(\cdot)$ functions (in the appendix). For the case with \textsc{10k} observations, we used
$K$$=$$50$ tasks with $N_k$$=$$200$ data-points and $M_k$$=$$3$
inducing variables in the sparse GP. The scenario for \textsc{100k} is
similar but divided into $K$$=$$250$ tasks with
$N_k$$=$$400$. Our method obtains better results than the exact distributed solutions due to the ensemble bound searches the average solution among all recyclable GPs. The baseline methods are based on a combination of solutions, if one is bad-fitted, it has a direct effect on the predictive performance. We also tested the data with the inference setup of \citet{gal2014distributed}, obtaining an \textsc{nlpd} of $2.58\pm0.11$ with $250$ nodes for \textsc{100k} data. It is better than ours and the baseline methods, but without a GP reconstruction, only distributes the computation of matrix terms. 
\textbf{(iii) Recyclable ensembles:} For a large synthetic
dataset ($N$$=$$10^{6}$), we tested the recyclable GPs
with $K$$=$$5\cdot10^{3}$ tasks as shown in Table \ref{tab:metrics}. However, if we ensemble large amount of local GPs, e.g. $K$$\gg$$10^{3}$, it is problematic for baseline methods, due to partitions must be revisited for building predictions and if one-of-many GP fails, performance decreases. Then, we repeated the experiment in a \textit{pyramidal} way. That is, building ensembles of recyclable ensembles, inspired in \citet{deisenroth2015distributed}. Our method obtained $\{\textsc{nlpd}$$=$$4.15, \textsc{rmse}$$=$$2.71$$, \textsc{mae}$$=$$2.27\}$. The results in Table \ref{tab:metrics} indicate that our model is more robust under the  \textit{concatenation} of approximations rather than overlapping them in the input space. % If the ensemble is repeated recursively, the error rates slightly augments compared with the 
%
%might be non efficient in memory. Then, we repeated the experiment in a \textit{pyramidal} way. That is, building ensembles of recyclable ensembles, inspired in the experiments of \citet{deisenroth2015distributed}. The results provided in Table \ref{tab:metrics} indicate that our model is more robust when \textit{concatenates} approximations rather than overlapping them in the input space. When the ensemble is repeated recursively, the error rates slightly augments compared with the exact inference solutions. 
%
%, obtaining the following metrics: $\{\textsc{nlpd}$$=$$4.15, \textsc{rmse}$$=$$2.71$$, \textsc{mae}$$=$$2.27\}$. 
%
%
\textbf{(iv) Solar physics dataset:} We tested the framework on solar
data (available at
  \texttt{https://solarscience.msfc.nasa.gov/}), which consists of
more than $N$$=$$10^{3}$ monthly average estimates of the sunspot
counting numbers from 1700 to 1995. We applied the mapping
$\log(1+y_i)$ to the output targets for performing Gaussian
regression. Metrics are provided in Table \ref{tab:solar}, where
std. values were small, so we do not include them. The perfomance with $50$ tasks is close to the baseline solutions, but without storing all distributed subsets of data.
\begin{table}[ht!]
	\vspace*{-2.0\baselineskip}
	\renewcommand{\thetable}{3}
	\caption{Comparative error metrics for distributed GP models}
	\label{tab:metrics}
	\centering
	\resizebox{\textwidth}{!}{%
		\begin{tabular}{cccccccccc}
			\toprule
			\textsc{Data size} $\rightarrow$ &&\textcolor{blue}{\textsc{10k}}&&&\textcolor{red}{\textsc{100k}}&&&\textsc{1m}&\\
			\midrule
			\textsc{Model}&\textcolor{blue}{\textsc{nlpd}}&\textcolor{blue}{\textsc{rmse}}&\textcolor{blue}{\textsc{mae}}&\textcolor{red}{\textsc{nlpd}}&\textcolor{red}{\textsc{rmse}}&\textcolor{red}{\textsc{mae}}&\textsc{nlpd}&\textsc{rmse}&\textsc{mae}\\
			\midrule
			BCM&$2.99\pm0.94$&$11.94\pm18.89$&$2.05\pm1.31$&$3.51\pm0.73$&$2.33\pm0.96$&$1.34\pm1.03$&$\textsc{na}$&$9.56\pm14.87$&$1.19\pm0.86$\\
			PoE &$2.79\pm0.16$&$2.32\pm0.22$&$1.86\pm0.22$&$2.82\pm0.67$&$2.19\pm0.91$&$1.71\pm0.84$&$2.91\pm0.63$&$1.98\pm0.61$&$1.32\pm0.05$\\
			GPoE &$2.79\pm0.56$&$2.43\pm0.52$&$1.96\pm0.48$&$\textcolor{red}{\mathbf{2.73}\pm\mathbf{0.72}}$&$2.19\pm0.91$&$1.71\pm0.84$&$2.72\pm0.52$&$1.98\pm0.61$&$\mathbf{1.32}\pm\mathbf{0.05}$\\
			RBCM &$2.96\pm0.51$&$2.49\pm0.51$&$2.02\pm0.46$&$3.03\pm0.86$&$2.51\pm1.12$&$1.99\pm1.04$&$\mathbf{2.56}\pm\mathbf{0.06}$&$\mathbf{1.82}\pm\mathbf{0.02}$&$1.37\pm0.03$\\
			%DVIGP &$\pm$&$\pm$&$\pm$&$\pm$&$\pm$&$\pm$&$\pm$&$\pm$&$\pm$\\
			\textbf{This work} &$\textcolor{blue}{\mathbf{2.71}\pm\mathbf{0.11}}$&$\textcolor{blue}{\mathbf{1.56}\pm\mathbf{0.04}}$&$\textcolor{blue}{\mathbf{0.97}\pm\mathbf{0.05}}$&$2.89\pm0.07$&$\textcolor{red}{\mathbf{1.73}\pm\mathbf{0.01}}$&$\textcolor{red}{\mathbf{1.23}\pm\mathbf{0.02}}$&$2.87\pm0.09$&$1.87\pm0.07$&$1.34\pm0.09$\\
			\bottomrule
	\end{tabular}}
	{\raggedright \scriptsize \textbf{Acronyms:} BCM \citep{tresp2000bayesian}, PoE \citep{ng2014hierarchical}, GPoE \citep{cao2014generalized} and RBCM \citep{deisenroth2015distributed}.\par}
	%{\raggedright \scriptsize RBCM \citep{deisenroth2015distributed} and DVIGP \citep{gal2014distributed}.\par}
	\vspace*{-\baselineskip}
\end{table}

\textbf{4.2~~~Classification.~} We adapted the entire recyclable GP
framework to accept non-Gaussian likelihoods, and in particular,
binary classification with Bernoulli distributions. We use the
\textit{sigmoid} mapping to link the GP function and the probit
parameters. 
\textbf{(iv) Pixel-wise MNIST classification:} Inspired in
the MNIST $\{0,1\}$ experiments of \citet{van2017convolutional}, we
threshold images of zeros and ones to black and white pixels. Then, to
simulate a pixel-wise learning scenario, we used each pixel as an
input-output datum whose input $\xc_{i}$ contains the two coordinates
($x_1$ and $x_2$ axes). Plots in Figure 4 illustrate that a predictive
ensemble can be built from smaller pieces of GP models, four corners
in the case of the number \textit{zero} and two for the
number \textit{one}. 
\textbf{(v) Compositional number:} As an illustration of potential applications of the recyclable GP approach, we build a number \textit{eight} predictor using exclusively two subsets of the approximations learned in the previous experiment with the image of a zero. The trick is to shift $\bm{Z}_k$ to place the local approximations in the desired position.
\textbf{(vi) Banana dataset:} We used the popular dataset in sparse GP classification for testing our method with $M$$=$$25$. We obtained a test \textsc{nlpd}$=7.21\pm0.04$, while the baseline variational GP test \textsc{nlpd} was $7.29\pm7.85$$\times$$10^{-4}$. The improvement is understandable as the total number of inducing points, including the local ones, is higher in the recyclable GP scenario. 
%\vspace*{-\baselineskip}
\begin{figure}[ht!]
	\label{fig:binary}
	\includegraphics[width=1.0\linewidth]{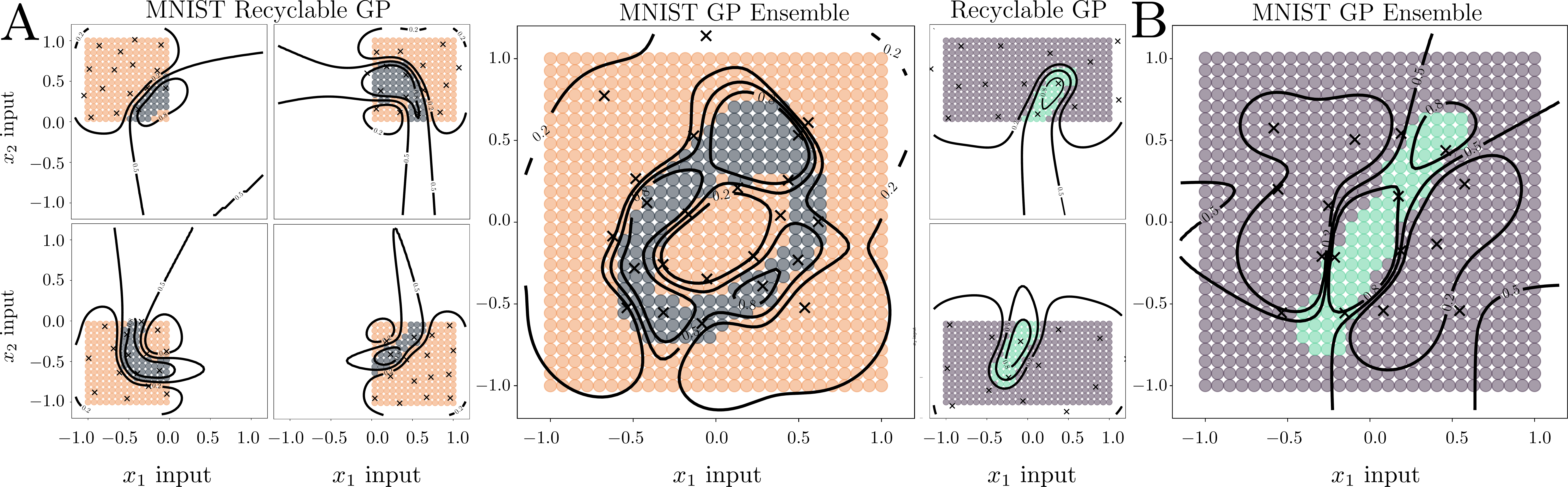}\vspace*{0.25cm}\\
	\vspace*{-\baselineskip}
	\begin{minipage}[b]{0.65\linewidth}
		\includegraphics[width=\linewidth]{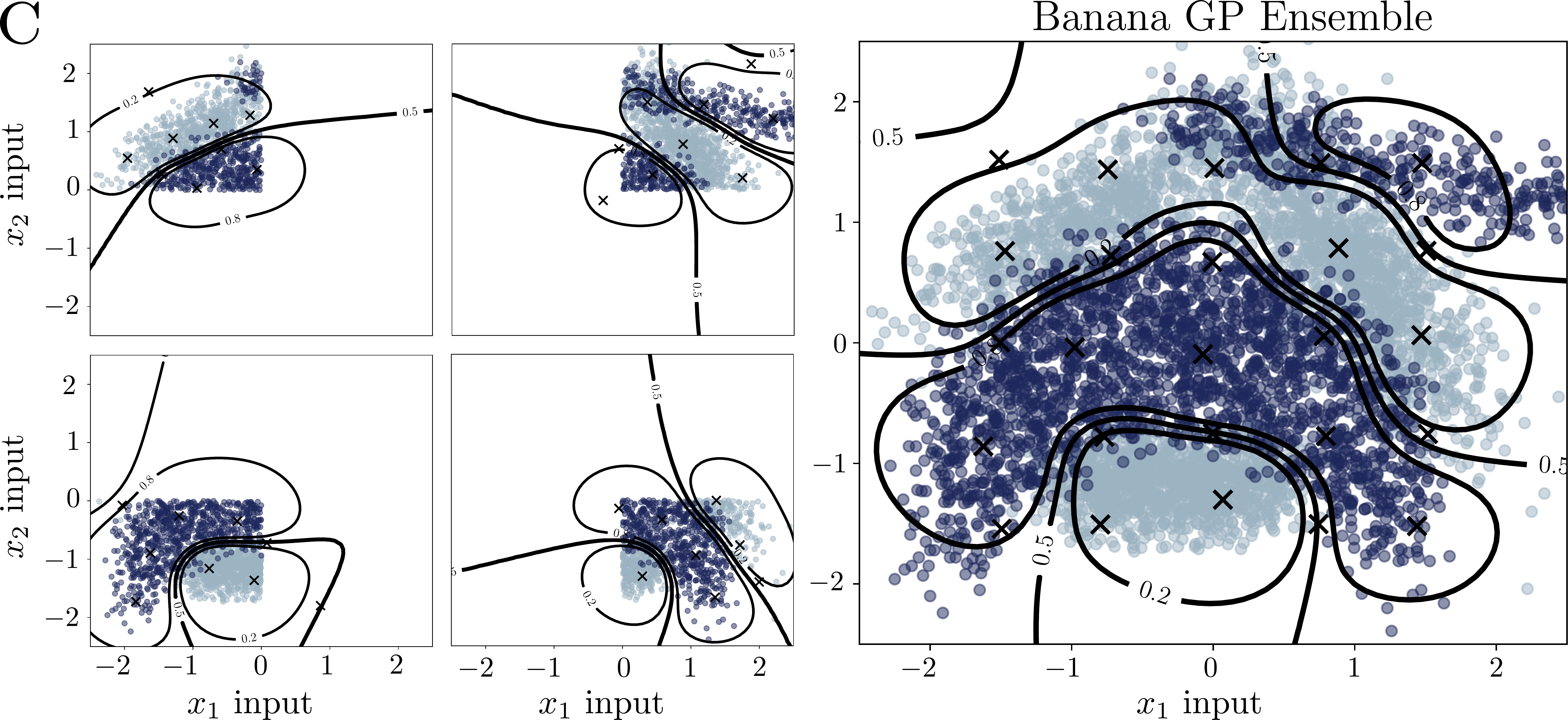}
		\caption{Recyclable GPs for $\{0,1\}$ MNIST \textbf{(A-B)}, banana \textbf{(C)} and compositional number \textit{eight} \textbf{(D)} experiments.}
		%\vspace*{0.4cm}
	\end{minipage}
	\begin{minipage}[b]{0.34\linewidth}
	\includegraphics[width=\linewidth]{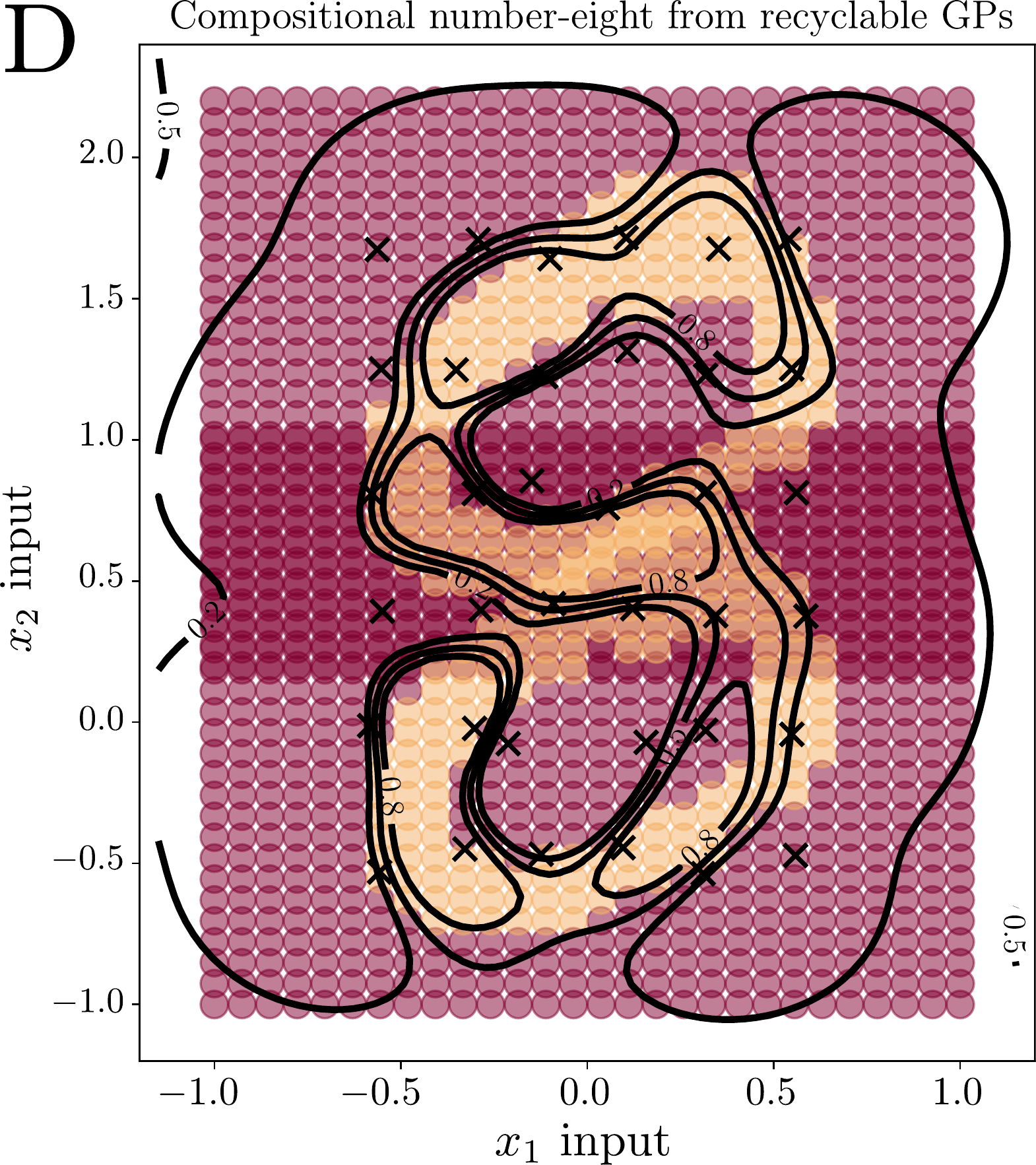}
	\end{minipage}
\end{figure}
\vspace*{0.25cm}
\begin{wrapfigure}[14]{l}{0.5\textwidth}
	\centering
	\vspace{-0.5cm}%
	\hspace{-3mm}%
	\includegraphics[width=1.0\linewidth]{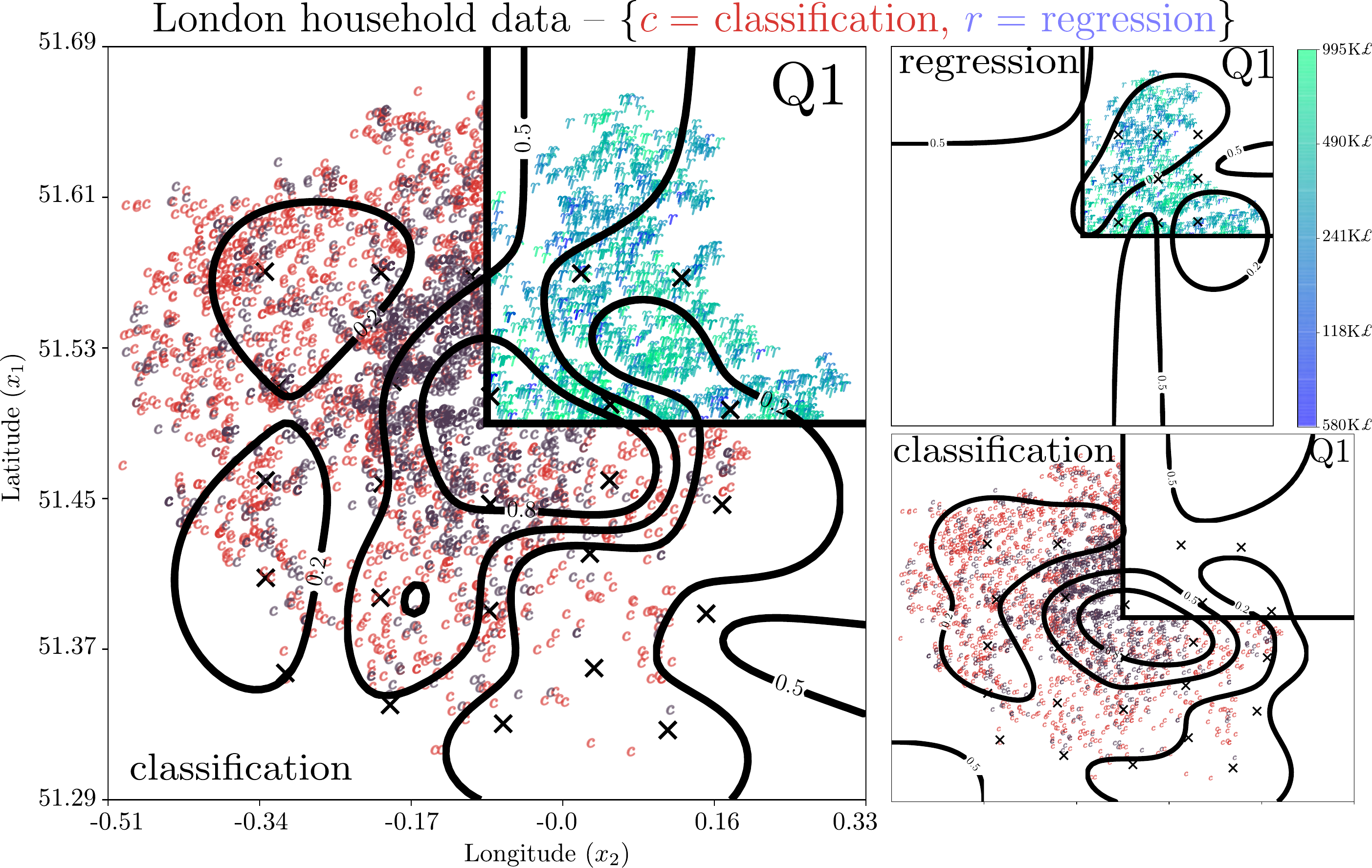}
	%\caption{Compositional eight-number recognition with the 4 recyclable GPs of the zero-number.}
	%\label{fig:graphic}
\end{wrapfigure}
\textbf{4.3~~~Heterogeneous tasks.~} We analysed how ensembles of recyclable GPs can be used if
one of the local tasks is regression and the other a GP
classifier. \textbf{(viii) London household data:} We have two
subsets of input-output variables: the binary contract of houses
(leasehold vs. freehold) and the price per latitude-longitude
coordinate in the London area. Three quadrants (Q) of the city $\{$Q2, Q3, Q4$\}$ are trained with a GP classifier and Q1 as regression. To clarify, Q1 is the right-upper corner given the central axes. Our purpose is to combine the local latent $\uc_k$, learned with the binary data on $\{$Q2, Q3, Q4$\}$ and the $\uc_k$ learned on Q1 via regression. 
Then, we search the global $f$ to be predict with a Bernoulli likelihood in Q1. %We want to asses whether the global f, learned in part with data from house prices on the other quadrants, can improve the prediction over el local model learned only with data from Q1. 
The ensemble shows a test \textsc{nlpd} of $7.94\pm0.01$ in classification while the recyclable task predicts with an \textsc{nlpd} of $8.00\pm0.01$ in the Q1. We asses that the heterogeneous GP prediction is better in Q1 than the local GP classifier. The mean GP of regression is passed through the sigmoid function to show the multimodality.
\vspace*{-0.75\baselineskip}
\section{Conclusion}
\vspace*{-0.5\baselineskip}

We introduced a novel framework for building global approximations from already fitted GP models. Our main contribution is the construction of ensemble bounds that accept parameters from regression, classification and heterogeneous GPs with different complexity without revisiting any data. We analysed its performance on synthetic and real data with different structure and likelihood models. Experimental results show evidence that the method is robust. In future work, it would be interesting to extend the framework to include convolutional kernels \citep{van2017convolutional} for image processing and functional regularisation \citep{titsias2020functional} for continual learning applications.

%\textbf{Broader Impact Statement.~} We identify two points of attention in the ethical aspects and future societal consequences of our recyclable GP framework. If one memorises and recycles functional models as GPs, old algorithms with biases could be repurposed for public applications. This has to be taken into account. We also argue that in the long-term, fitted models must be protected, as we do with data, for preserving the privacy of individuals.

\subsection*{Acknowledgements}

This work was supported by the Ministerio de Ciencia, Innovaci{\'o}n y Universidades under grant TEC2017-92552-EXP (aMBITION), by the Ministerio de Ciencia, Innovaci{\'o}n y Universidades, jointly with the European Commission (ERDF), under grants TEC2017-86921-C2-2-R (CAIMAN) and RTI2018-099655-B-I00 (CLARA), and by the Comunidad de Madrid under grant Y2018/TCS-4705 (PRACTICO-CM). PMM acknowledges the support of his FPI grant BES-2016-077626 from the Ministerio of Econom\'ia of Spain. MAA has been financed by the EPSRC Research Projects EP/R034303/1 and EP/T00343X/1. MAA has also been supported by the Rosetrees Trust (ref: A2501).

%ethical aspects and future societal consequences   ()

 %fAn extended idea of our paper would be  

\small
\bibliography{neurips2020}
%\nocite{*}

\newpage
\appendix
\subsection*{Appendix A. ~Detailed Derivation of the Lower Ensemble Bound}

The construction of ensemble variational bounds from recyclable GP models is based on the idea of \textit{augmenting} the marginal likelihood to be conditioned on the infinite-dimensional GP function $\finf$. Notice that $\finf$ contains \underline{all} the function values taken by $f(\cdot)$ over the input-space $\mathbb{R}^{p}$, including the input targets $\{\xc_i\}^{N}_{i=1}$, the local \textit{inducing-inputs} $\{\bm{Z}_{k}\}^{K}_{k=1}$ and the global ones $\bm{Z}_{*}$. Having $K$ partitions of the dataset $\Dcal$ with their corresponding outputs $\yc = \{\yc_1, \yc_2, \dots, \yc_K\}$, we begin by augmenting the marginal log-likelihood as
\begin{equation}
	\label{eq:marginal}
	\log p(\yc) = \log p(\yc_1, \yc_2, \dots, \yc_K) = \log \int p(\yc, f_{\infty})df_{\infty}, 
\end{equation}
that factorises according to
\begin{equation}
	\label{eq:factorising}
	\log \int p(\yc, f_{\infty})df_{\infty} = \log \int p(\yc| f_{\infty})p(f_{\infty})df_{\infty},
\end{equation}
where $p(\yc| f_{\infty})$ is the \textit{augmented} likelihood term of all the output targets of interest and $p(f_{\infty})$ the GP prior over the infinite amount of points in the input-space $\mathbb{R}^{p}$. This last distribution takes the form of an infinite-dimensional Gaussian, that we avoid to evaluate explicitly in the equations. To build the lower bound on the log-marginal likelihood, we first introduce the global variational distribution $q(\bm{u_*}) = \mathcal{N}(\bm{u_*}|\bm{\mu_*}, \bm{S_*})$ into the equation,
\begin{multline}
	\log p(\yc) = \log \int p(\yc| f_{\infty})p(f_{\infty})df_{\infty} = \log \int \frac{q(\up) }{q(\up) }p(\yc| f_{\infty})p(f_{\infty})df_{\infty}\\
	=\log \iint \frac{q(\up)}{q(\up)}p(\yc|\finf)p(f_{\infty\neq\up}|\up)p(\up)df_{\infty\neq\up}d\up.
\end{multline}
Notice that the differentials $df_{\infty}$ have been splitted into $df_{\infty\neq\up}d\up$, and at the same time, we applied properties of Gaussian conditionals in the GP prior to rewrite $p(f_{\infty})$ as 
$p(f_{\infty\neq\up}|\up)p(\up)$. When the target variables $\up$ are explicit in the expression, our second step is the application of the Jensen inequality twice as it is done in the reparameterisation of \citep{gal2014distributed}, that is

\begin{multline}
	\log p(\yc) = \log \iint \frac{q(\up)}{q(\up)}p(\yc|\finf)p(f_{\infty\neq\up}|\up)p(\up)df_{\infty\neq\up}d\up\\
	=\log \iint q(\up)p(f_{\infty\neq\up}|\up)p(\yc|\finf)\frac{p(\up)}{q(\up)}df_{\infty\neq\up}d\up\\
	= \log \left(\mathbb{E}_{q(\uc_{*})}\left[\mathbb{E}_{p(\fc_{\infty\neq\uc_*}|\uc_*)}\left[ p(\yc|\finf)\frac{p(\up)}{q(\up)} \right] \right]\right)\\
	\geq \mathbb{E}_{q(\uc_{*})}\left[\log \left(\mathbb{E}_{p(\fc_{\infty\neq\uc_*}|\uc_*)}\left[ p(\yc|\finf)\frac{p(\up)}{q(\up)} \right]\right) \right]\\
	\geq \mathbb{E}_{q(\uc_{*})}\left[\mathbb{E}_{p(\fc_{\infty\neq\uc_*}|\uc_*)}\left[\log \left( p(\yc|\finf)\frac{p(\up)}{q(\up)}\right)\right] \right] = \Lcal_{\mathcal{E}}.
	\label{eq:init_bound}
\end{multline}

Then, if we have \eqref{eq:init_bound}, which is the first version of our ensemble lower bound $\Lcal_{\mathcal{E}}$, we can use the augmented likelihood term $p(\yc|\finf)$ to introduce the local approximations to $f$ instead of revisiting the data. This is,
\begin{multline}
	\Lcal_{\mathcal{E}} = \mathbb{E}_{q(\uc_{*})}\left[\mathbb{E}_{p(\fc_{\infty\neq\uc_*}|\uc_*)}\left[  \log p(\yc|\finf) + \log \left(\frac{p(\up)}{q(\up)}\right)\right] \right]\\
	= \mathbb{E}_{q(\uc_{*})}\left[\mathbb{E}_{p(\fc_{\infty\neq\uc_*}|\uc_*)}\left[  \log p(\yc|\finf) \right]  - \log \left(\frac{q(\up)}{p(\up)}\right)\right]\\
	= \mathbb{E}_{q(\uc_{*})}\left[\mathbb{E}_{p(\fc_{\infty\neq\uc_*}|\uc_*)}\left[  \sum^{K}_{k=1}\log p(\yc_k|\finf) \right]  - \log \left(\frac{q(\up)}{p(\up)}\right)\right]\\
	= \mathbb{E}_{q(\uc_{*})}\left[\sum^{K}_{k=1} \mathbb{E}_{p(\fc_{\infty\neq\uc_*}|\uc_*)}\left[ \log p(\yc_k|\finf) \right]  - \log \left(\frac{q(\up)}{p(\up)}\right)\right],
	\label{eq:bound_k}
\end{multline}
where the log-ratio $q(\up)/p(\up)$ acts as a constant to the second expectation $\mathbb{E}_{p(\fc_{\infty\neq\uc_*}|\uc_*)}\left[\cdot \right]$ and we applied \textit{conditional independence} (CI) among all the output partitions given the latent function $\finf$. That is, we introduced $p(\yc|\finf) = \prod^{K}_{k=1}p(\yc_k|\finf)$ to factorise the expectation term in \eqref{eq:bound_k} across the $K$ tasks.

Under the approximation of $p(\yc_k|\finf)$ obtained by inverting the Bayes theorem, we use $p(\yc_k|\finf) \approx q_k(\finf)/ p_k(\finf)$ to introduce the local posterior distributions $q_k(\cdot)$ and priors $p_k(\cdot)$ in the bound $\Lcal_{\mathcal{E}}$. This leads to

\begin{multline}
	\Lcal_{\mathcal{E}}  = \mathbb{E}_{q(\uc_{*})}\left[\sum^{K}_{k=1} \mathbb{E}_{p(\fc_{\infty\neq\uc_*}|\uc_*)}\left[ \log p(\yc_k|\finf) \right]  - \log \left(\frac{q(\up)}{p(\up)}\right)\right]\\
	\approx \mathbb{E}_{q(\uc_{*})}\left[\sum^{K}_{k=1} \mathbb{E}_{p(\fc_{\infty\neq\uc_*}|\uc_*)}\left[ \log \left(\frac{q_k(\finf)}{p_k(\finf)}\right) \right]  - \log \left(\frac{q(\up)}{p(\up)}\right)\right]\\
	= \mathbb{E}_{q(\uc_{*})}\left[\sum^{K}_{k=1} \mathbb{E}_{p(\fc_{\infty\neq\uc_*}|\uc_*)}\left[ \log \left(\frac{\cancel{p(\fc_{\infty\neq \uc_k}|\uc_k)}q_k(\uk)}{\cancel{p(\fc_{\infty\neq \uc_k}|\uc_k)}p_k(\uk)}\right) \right]  - \log \left(\frac{q(\up)}{p(\up)}\right)\right]\\
	= \mathbb{E}_{q(\uc_{*})}\left[\sum^{K}_{k=1} \mathbb{E}_{p(\fc_{\infty\neq\uc_*}|\uc_*)}\left[ \log \left(\frac{q_k(\uk)}{p_k(\uk)}\right) \right]  - \log \left(\frac{q(\up)}{p(\up)}\right)\right],
	\label{eq:bound}
\end{multline}
where we now have the explicit local distributions $q_k(\uk)$ and $p_k(\uk)$ on the subsets of inducing-inputs $\{\bm{Z}_k\}^{K}_{k=1}$. The cancellation of conditionals is a result of the variational factorization \citep{titsias2009variational}. Looking to the last version of the bound in \eqref{eq:bound}, there is still one point that maintains the infinite-dimensionality, the conditional prior  $p(\fc_{\infty\neq\uc_*}|\uc_*)$ and its corresponding expectation term $\mathbb{E}_{p(\fc_{\infty\neq\uc_*}|\uc_*)}\left[\cdot \right]$. To adapt it to the local inducing variables $\uk$, we apply the following simplification to each $k$-th integral in \eqref{eq:bound} based in the properties of Gaussian marginals (see section A.1),
\begin{multline}
	\mathbb{E}_{p(\fc_{\infty\neq\uc_*}|\uc_*)}\left[ \log \left(\frac{q_k(\uk)}{p_k(\uk)}\right) \right] = \int p(\fc_{\infty\neq\uc_*}|\uc_*)\log \left(\frac{q_k(\uk)}{p_k(\uk)}\right) d\fc_{\infty\neq\uc_*}\\
	= \iint p(\fc_{\infty\neq\{\uc_*,\uk\}}, \uk |\uc_*)\log \left(\frac{q_k(\uk)}{p_k(\uk)}\right) d\fc_{\infty\neq\{\uc_*,\uk\}}d\uk\\
	= \int p(\uk |\uc_*)\log \left(\frac{q_k(\uk)}{p_k(\uk)}\right) d\uk = \mathbb{E}_{p(\uk|\uc_*)}\left[ \log \left(\frac{q_k(\uk)}{p_k(\uk)}\right) \right].
\end{multline}
This is the expectation that we plug in the final version of the bound, to obtain
\begin{multline}
	\Lcal_{\mathcal{E}} = \mathbb{E}_{q(\uc_{*})}\left[\sum^{K}_{k=1} \mathbb{E}_{p(\uk|\uc_*)}\left[ \log \left(\frac{q_k(\uk)}{p_k(\uk)}\right) \right]  - \log \left(\frac{q(\up)}{p(\up)}\right)\right]\\
	= \sum^{K}_{k=1}\mathbb{E}_{q(\uc_{*})}\left[ \mathbb{E}_{p(\uk|\uc_*)}\left[ \log \left(\frac{q_k(\uk)}{p_k(\uk)}\right) \right] \right]  - \mathbb{E}_{q(\uc_{*})}\left[\log \left(\frac{q(\up)}{p(\up)}\right)\right]\\
	= \sum^{K}_{k=1}\mathbb{E}_{q(\uc_{*})}\left[ \mathbb{E}_{p(\uk|\uc_*)}\left[ \log \left(\frac{q_k(\uk)}{p_k(\uk)}\right) \right] \right]  - \text{KL}\left[q(\up)||p(\up)\right]\\
	= \sum^{K}_{k=1}\mathbb{E}_{q_{\mathcal{C}}(\uk)}\left[\log q_k(\uk) - \log p_k(\uk) \right]   - \text{KL}\left[q(\up)||p(\up)\right],
	\label{eq:ensemble_bound}
\end{multline}
where $q_{\mathcal{C}}(\uk)$ is the \textit{contrastive} predictive GP posterior, whose derivation is provided in the section A.2. Importantly, the ensemble bound in \eqref{eq:ensemble_bound} is the one that we aim to maximise w.r.t. some variational parameters and hyperparameters. For a better comprehension of this point, we provide an extra-view of the bound and the presence of (fixed) local and (unfixed) global parameters in each term. See section A.3. for this.
\subsubsection*{Appendix A.1. ~Gaussian marginals for infinite-dimensional integral operators}

The properties of Gaussian marginal distributions indicate that, having two \textit{normal}-distributed random variables $\bm{a}$ and $\bm{b}$, its joint probability distribution is given by 
$$p(\bm{a}, \bm{b}) = \Ncal\left(\begin{bmatrix}
\bm{\mu}_a\\\bm{\mu}_b\end{bmatrix}, \begin{bmatrix}\bm{\Sigma}_{aa}&\bm{\Sigma}_{ab}\\\bm{\Sigma}_{ba}&\bm{\Sigma}_{bb}\end{bmatrix}\right),$$
and if we want to marginalize one of that variables out, such as $\int p(\bm{a}, \bm{b}) d\bm{b}$. It turns to be
$$\int p(\bm{a}, \bm{b}) d\bm{b} = p(\bm{a}) = \Ncal(\bm{\mu}_a, \bm{\Sigma}_{aa}).$$
This same property is applicable to every derivation with GPs. In our case, it is the key point that we use to reduce the infinite-dimensional integral operators over the full stochastic processes. An example can be found in the expectation $\mathbb{E}_{p(\fc_{\infty\neq\uc_*}|\uc_*)}\left[\cdot \right]$ of \eqref{eq:bound}. Its final derivation to only integrate on $\uk$ rather than on $\fc_{\infty\neq\uc_*}$ comes from
\begin{multline*}
	p(\fc_{\infty\neq\uc_*}|\uc_*) = p(\fc_{\infty\neq\{\up, \uk\}}, \uk|\up)\\
	= \Ncal\left(\begin{bmatrix}
		\bm{m}_{\fc_{\infty\neq\{\up, \uk\}}|\up}\\\bm{m}_{\uk|\up}\end{bmatrix}, \begin{bmatrix}\bm{Q}_{\fc_{\infty\neq\{\up, \uk\}}|\up }&\bm{Q}_{\fc_{\infty\neq\{\up, \uk\}},\uk|\up}\\\bm{Q}_{\uk, \fc_{\infty\neq\{\up, \uk\}}|\up}&\bm{Q}_{\uk|\up}\end{bmatrix}\right),
\end{multline*}
and if we marginalize over $\fc_{\infty\neq\{\up, \uk\}}|\up$, ends in the following reduction of the conditional prior expectation
\begin{multline}
	\mathbb{E}_{p(\fc_{\infty\neq\uc_*}|\uc_*)}\left[ g(\uk) \right] = \int p(\fc_{\infty\neq\uc_*}|\uc_*)g(\uk) d\fc_{\infty\neq\uc_*}\\
	= \iint p(\fc_{\infty\neq\{\uc_*,\uk\}}, \uk|\uc_*)g(\uk) d\fc_{\infty\neq\{\uc_*,\uk\}}d\uk\\
	= \int p(\uk|\uc_*)g(\uk)d\uk = \mathbb{E}_{p(\uk|\uc_*)}\left[ g(\uk) \right],
\end{multline}
where we denote $g(\uk) = \log \left( q_k(\uk)/p_k(\uk)\right)$ and we used
$$\int p(\fc_{\infty\neq\{\up, \uk\}}, \uk|\up) d\fc_{\infty\neq\{\uc_*,\uk\}} = p(\uk) = \Ncal(\bm{m}_{\uk|\up}, \bm{Q}_{\uk|\up}).$$
\subsubsection*{Appendix A.2. ~Contrastive posterior GP predictive}
The \textit{contrastive} predictive GP posterior distribution $q_{\mathcal{C}}(\uk)$ is obtained from the \textit{nested} integration in \eqref{eq:ensemble_bound}. We begin its derivation with the l.h.s. expectation term in \eqref{eq:ensemble_bound}, then
\begin{multline}
	\sum^{K}_{k=1}\mathbb{E}_{q(\up)}\left[ \mathbb{E}_{p(\uk|\up)}\left[ \log \left(\frac{q_k(\uk)}{p_k(\uk)}\right) \right] \right]\\
	= \sum^{K}_{k=1}\iint q(\up)p(\uk|\up)\log \left(\frac{q_k(\uk)}{p_k(\uk)}\right)d\uk d\up\\
	= \sum^{K}_{k=1}\int \underset{q_{\mathcal{C}}(\uk)}{\underline{\left( \int q(\up)p(\uk|\up) d\up \right)}} \log \left(\frac{q_k(\uk)}{p_k(\uk)}\right)d\uk,
	\label{eq:contrastive}	
\end{multline}
where the conditional GP prior distribution between the local inducing-inputs $\uk$ and the global ones $\up$, is $p(\uc_k|\up) = \mathcal{N}(\uc_k|\bm{m}_{k|*}, \bm{Q}_{k|*})$ with
\begin{align*}
	\bm{m}_{k|*} &= \Kpk^\top\Kpp^{-1}\up,\\
	\bm{Q}_{k|*} &= \Kk - \Kpk^\top\Kpp^{-1}\Kpk,
\end{align*}
and where covariance matrices are built from $\left[\Kpp\right]_{m,n} := k(\zc_m, \zc_n)$ with $\zc_m, \zc_n \in \mathbb{R}^{p}$. Finally, the contrastive predictive GP posterior $q_{\mathcal{C}}(\uk)$ can be computed from the expectation term in \eqref{eq:contrastive} as
\begin{equation}
	\int q(\up)p(\uk|\up) d\up = q_{\mathcal{C}}(\uk) = \Ncal(\uk| \bm{m}_{\mathcal{C}}, \bm{S}_{\mathcal{C}}), 
\end{equation}
where the parameters $\bm{m}_{\mathcal{C}}$ and $\bm{S}_{\mathcal{C}}$ are
\begin{align*}
	\bm{m}_\mathcal{C} &= \Kpk^\top\Kpp^{-1}\mup,\\
	\bm{S}_\mathcal{C} &= \Kk - \Kpk^\top\Kpp^{-1}(\Sp - \Kpp)\Kpp^{-1}\Kpk.
\end{align*}
\subsubsection*{Appendix A.3. ~Parameters in the lower ensemble bound}

We approximate the global approximation to the GP posterior distribution as $q(f) \approx p(f|\Dcal)$. Additionally, we introduce the subset of global inducing-inputs $\bm{Z_{*}} = \{\zc_m\}^{M}_{m=1}$ and their corresponding function evaluations are $\up$. Then, the \textit{explicit} variational distribution given the pseudo-observations $\up$ is $q(\up) = \Ncal(\up| \mup, \Sp)$. Previously, we have obtained the list of objects $\Ecal = \{\Ecal_1, \Ecal_2, \dots, \Ecal_K\}$ without any specific order, where each $\Ecal_k = \{\bm{\phi_k}, \bm{\psi_k}, \bm{Z}_k\}$, $\bm{\phi_k}$ being the corresponding local variational parameters $\bm{\mu}_k$ and $\bm{S}_k$.

If we look to the ensemble lower bound in \eqref{eq:ensemble_bound}, we omitted the conditioning on both variational parameters and hyperparameters for clarity. However, to make this point clear, we will now rewritte \eqref{eq:ensemble_bound} to show the influence of each parameter variable over each term in the global bound. We remark that \textcolor{red}{$\{\bm{\phi_k}, \bm{\psi_k}\}^{K}_{k=1}$} are given and \underline{fixed}, whilst 
\textcolor{blue}{$\{\bm{\phi_{*}}, \bm{\psi_{*}}\}$} are the variational parameters and hyperparameters that we aim to fit,
\begin{equation*}
	\Lcal_{\mathcal{E}}(\textcolor{blue}{\bm{\phi_{*}}, \bm{\psi_{*}}}) = \sum^{K}_{k=1}\mathbb{E}_{q_{\mathcal{C}}(\uk|\textcolor{blue}{\bm{\phi_{*}, \psi_{*}}})}\left[\log q_k(\uk|\textcolor{red}{\bm{\phi_k}}) - \log p_k(\uk|\textcolor{red}{\bm{\psi_k}}) \right]   - \text{KL}\left[q(\up|\textcolor{blue}{\bm{\phi_{*}}})||p(\up|\textcolor{blue}{\bm{\psi_{*}}})\right].
\end{equation*}
We remind that the global variational parameters are \textcolor{blue}{$\bm{\phi_{*}} = \{\mup, \Sp\}$}, while the hyperparameters would correspond to \textcolor{blue}{$\bm{\psi_{*}} = \{\ell, \sigma_a\}$} in the case of using the vanilla \textit{kernel}, with $\ell$ being the lengthscale and $\sigma_a$ the amplitude variables. The notation of the local counterpart is equivalent.

The dependencies of parameters in our Pytorch implementation (\url{https://github.com/pmorenoz/RecyclableGP}) are clearly shown and evident from the code structure oriented to objects. It is also amenable for the introduction of new covariance functions and more structured variational approximations if needed.
%\subsubsection{Average global GP solutions}

\subsection*{Appendix B. ~Distributions and Expectations}

To assure the future and easy reproducibility of our recyclable GP framework, we provide the exact expression of all distributions and expectations involved in the lower ensemble bound in \eqref{eq:ensemble_bound}. 

\noindent\textbf{Distributions:~} The log-distributions and distributions that appear in \eqref{eq:ensemble_bound} are $\log q(\uc_k)$, $\log p(\uc_k)$, $q(\up)$, $p(\up)$ and $q_{\mathcal{C}}(\uk)$. First, the computation of the logarithmic distributions is
$$\log q(\uc_k) = \log \left(\mathcal{N}(\uc_k|\muk, \Sk)\right) =  -\frac{1}{2} (\uc_k - \muk)^\top \Sk^{-1}(\uc_k - \muk)  -\frac{1}{2}\log\det(2\pi\Sk),$$

$$\log p(\uc_k) = \log \left(\mathcal{N}(\uc_k|\bm{0}, \K_{kk})\right) =  -\frac{1}{2} \uc_k^\top \K_{kk}^{-1}\uc_k  -\frac{1}{2}\log\det(2\pi\K_{kk}),$$
while $q(\up)$ and $p(\up)$ are just $q(\up) = \mathcal{N}(\up|	\mup, \Sp)$ and $p(\up) = \mathcal{N}(\up|	\bm{0}, \Kpp)$. The exact expression of the distribution $q_{\mathcal{C}}(\uk)$ is provided in the section A.2.

\noindent\textbf{Expectations and divergences:} The $K$ expectations in the l.h.s.\ term in \eqref{eq:ensemble_bound} can be rewritten as
\begin{multline}
	\sum^{K}_{k=1}\mathbb{E}_{q_{\mathcal{C}}(\uk)}\left[\log q_k(\uk) - \log p_k(\uk) \right]\\ = \sum^{K}_{k=1}\left[\mathbb{E}_{q_{\mathcal{C}}(\uk)}\left[\log q_k(\uk)\right] - \mathbb{E}_{q_{\mathcal{C}}(\uk)}\left[\log p_k(\uk) \right]\right]\\
	= \sum^{K}_{k=1}\left[\Big\langle\log q_k(\uk)\Big\rangle_{q_{\mathcal{C}}(\uk)} - \Big\langle\log p_k(\uk) \Big\rangle_{q_{\mathcal{C}}(\uk)}\right],
\end{multline}
where the $k$-th expectations over both $\log q_k(\uk)$ and $\log p_k(\uk)$ take the form
$$\Big\langle\log q_k(\uk)\Big\rangle_{q_{\mathcal{C}}(\uk)} = -\frac{1}{2}\left(\text{Tr}\left(\Sk^{-1}\bm{S}_\mathcal{C} \right) + (\bm{m}_\mathcal{C} - \muk)^{\top}\Sk^{-1}(\bm{m}_\mathcal{C} - \muk) + \log\det\left(2\pi\Sk\right)\right),$$
$$\Big\langle\log p_k(\uk)\Big\rangle_{q_{\mathcal{C}}(\uk)} = -\frac{1}{2}\left(\text{Tr}\left(\K_{kk}^{-1}\bm{S}_\mathcal{C} \right) + \bm{m}_\mathcal{C}^{\top}\K_{kk}^{-1}\bm{m}_\mathcal{C} + \log\det\left(2\pi\K_{kk}\right)\right).$$

\subsection*{Appendix C. ~Combined Ensemble Bounds with Unseen Data}

As we already mentioned in the manuscript,  there might be scenarios where it could be not necessary to distribute the whole dataset $\Dcal$ in $K$ local tasks or, for instance, a new \textit{unseen} subset $k+1$ of observations might be available for processing. In such case, it is still possible to obtain a \textit{combined} global solution that fits both to the local GP approximations and the new data. For clarity on this point, we rewrite the principal steps of the ensemble bound derivation in section A but without substituting all the log-likelihood terms by its Bayesian approximation, that is
\begin{multline}
	\Lcal_{\mathcal{E}}  = \mathbb{E}_{q(\uc_{*})}\left[\mathbb{E}_{p(\fc_{\infty\neq\uc_*}|\uc_*)}\left[ \sum^{K}_{k=1}\log p(\yc_k|\finf) + \log p(\yc_{k+1}|\finf) \right]  - \log \left(\frac{q(\up)}{p(\up)}\right)\right]\hspace{1cm}\\
	= \mathbb{E}_{q(\uc_{*})}\left[\sum^{K}_{k=1} \mathbb{E}_{p(\fc_{\infty\neq\uc_*}|\uc_*)}\left[ \log p(\yc_k|\finf)\right] + \mathbb{E}_{p(\fc_{\infty\neq\uc_*}|\uc_*)}\left[ \log p(\yc_{k+1}|\finf) \right]  - \log \left(\frac{q(\up)}{p(\up)}\right)\right]\\
	= \mathbb{E}_{q(\uc_{*})}\left[\sum^{K}_{k=1} \mathbb{E}_{p(\uk|\uc_*)}\left[ \log \left(\frac{q_k(\uk)}{p_k(\uk)}\right)\right] + \mathbb{E}_{p(\bm{f}_{k+1}|\uc_*)}\left[ \log p(\yc_{k+1}|\bm{f}_{k+1}) \right]  - \log \left(\frac{q(\up)}{p(\up)}\right)\right]\\
	= \sum^{K}_{k=1}\mathbb{E}_{q_{\mathcal{C}}(\uk)}\left[\log q_k(\uk) - \log p_k(\uk) \right] + \sum_{i=1}^{N_{k+1}}\mathbb{E}_{q(\f_i)} \left[\log p(y_i|\f_i)\right]  - \text{KL}\left[q(\up)||p(\up)\right],
	\label{eq:combined}
\end{multline}
where $q(\f_i)$ is the result of the integral $q(\f_i) = \int q(\up)p(\f_i|\up) d\up$ and we applied the factorisation to the \textit{new} $(k+1)$-th expectation term as in \citet{hensman2015scalable}.

\subsection*{Appendix D. ~Intractable Expectations}

When we consider a binary classification task, the likelihood function use to be a Bernoulli distribution, such as $p(y_i|\f_i) = \text{Ber}(y_i|\rho = \phi(\f_i))$. The non-linear linking mapping $\phi(\cdot)$ is the \textit{sigmoid} function in our case. However, for training the local GP approximations, the expectation term of the ELBO is still intractable over the log-likelihood distribution. To solve the following integrals

$$\mathbb{E}_{q(\f_i)}\left[\log p(y_i|\f_i)\right] = \int q(\f_i) \log p(y_i|\f_i) d\f_i,$$

we make use of the Gaussian-Hermite quadratures. In the univariate case with binary observations, the previous integral can be approximated as

$$\mathbb{E}_{q(\f_i)}\left[\log p(y_i|\f_i)\right] \approx \frac{1}{\sqrt{\pi}} \sum_{s=1}^{S} w_s \log p(y_i|\sqrt{2\bm{v}_i}\f_s + \bm{m}_i),$$
where $\bm{m}_i$ and $\bm{v}_i$ are the corresponding mean and variance of the marginal variational distribution $q(\f_i)$. Additionally, the pairs of weight-point values $(w_s, \f_s)$ are obtained by sampling $S$ times the Hermite polynomial $H_n(x) = (-1)^{n}e^{x^2}\frac{d^{n}}{dx^{n}}e^{-x^2}$. This computation is also used for the calculus of predictive distributions and \textsc{nlpd} metrics.

\subsection*{Appendix E. ~Experiments, Optimization Algorithms and Metrics}

The code for the experiments is written in Python 3.7 and uses the Pytorch syntax for the automatic differentiation of the probabilistic models. It can be found in the repository \url{https://github.com/pmorenoz/RecyclableGP}, where we also use the library GPy for some algebraic utilities. In this section, we provide a detailed description of the experiments and the data used, the initialization of both variational parameters and hyperparameters, the optimization algorithm for both the local and the global GP and the performance metrics included in the main manuscript, e.g. the negative log-predictive density (\textsc{nlpd}), the root mean square error (\textsc{rmse}) and  the mean absolute error (\textsc{mae}).

\subsubsection*{Appendix E.1. ~Detailed description of experiments}

In our experiments with toy data, we used two versions of the same sinusoidal function, one of them with an incremental bias. The true expressions of $f(\cdot)$ are

$$ f(x) = \frac{9}{2}\cos \left(2\pi x + \frac{3\pi}{2}\right) - 3\sin \left(\frac{43\pi}{10} x + \frac{3\pi}{10}\right),$$
and 
$$ f(x)_{\text{bias}} = f(x) + 3x - \frac{15}{2}.$$
\textbf{i) Toy concatenation:} For the first experiment, whose results are illustrated in the Figure 2 of the main manuscript, we generated $K=5$ subsets of observations in the input-space range $\xc \in [0.0, 5.5]$. Each subset was formed by $N_k = 500$ uniform samples of $\xc_k$  that were later evaluated by $f(x)_{\text{bias}}$. Having the values of the true underlying function $\fc_{k} = f(\xc_k)$, we generated the true output targets as $\yc_k = \fc_{k} + \epsilon_k$, where $\epsilon_k \sim \Ncal(0, 2)$. For each local task, we set a number of $M_k = 15$ inducing-inputs $\bm{Z}_k$ that were initially equally spaced in each local input region. The chosen number of global inducing-inputs $\bm{Z_{*}}$ was $M=35$, initialized in the same manner as in the local case. For all the posterior predictive GPs plotted, we used $N_\text{test} = 400$ also equally spaced in the global input-space. The setup of the VEM algorithm (see section E.3) was \{$\textsc{ve}=30$, $\textsc{vm}=10$, $\eta_{m} = 10^{-3}$, $\eta_{L} = 10^{-6}$, $\eta_{\psi} = 10^{-8}$, $\eta_{Z} = 10^{-8}$\} for the ensemble GP. The previous variables $\eta$ and \textsc{vm} refer to the learning rates used for each type of parameter and the number of iterations in the optimization algorithm. 

\textbf{ii) Distributed GPs:} In this second experiment, our goal is to compare the performance of the recyclable framework with the distributed GP methods in the literature \citep{tresp2000bayesian,ng2014hierarchical,cao2014generalized,deisenroth2015distributed}. To do so, we begin by generating toy samples from the sinusoidal function $f(x)$. The comparative experiment is divided in two parts, in one, we observe $N=10^{3}$ and in the other, $N=10^{4}$ input-output data points. In the first case, we splitted the dataset $\Dcal$ in $K=50$ tasks with $N_k=200$ and $M_k=3$ per partition. Any of these distributed subsets were overlapping, and their corresponding input-spaces concatenated perfectly in the range $\xc \in [0.0, 5.5]$. For the setting with $N=10^{4}$ samples, we used $K=500$ local tasks, that in this case, were overlapping. As we already commented in the main manuscript, the baseline methods underperform more than our framework in problems where partitions do not overlap in the input-space. Additionally, \textit{standard deviation} (std.) values in Table 3 indicate that we are more robust to the fitting crash of some task. This fact is understandable as our method searches a global solution $q(\up)$ that fits to all the local GPs in average. In contrast, the baseline methods are based on a final ensemble solution that is an analytical combination of all the distributed ones. Then, if one or more fails, the final predictive performance might be catastrophic. Notice that the baseline methods only require to train the local GPs separately, thing that we did with the LBFGS optimization algorithm. The setup of the VEM algorithm during the ensemble fitting was \{$\textsc{ve}=30$, $\textsc{vm}=10$, $\eta_{m} = 10^{-3}$, $\eta_{L} = 10^{-6}$, $\eta_{\psi} = 10^{-8}$, $\eta_{Z} = 10^{-8}$\}. As in the previous experiment with toy data, we set $M=35$ inducing-inputs. 

\begin{wrapfigure}[22]{r}{0.5\textwidth}
	\centering
	\vspace{-0.4cm}%
	\hspace{+5mm}%
	\includegraphics[width=1.0\linewidth]{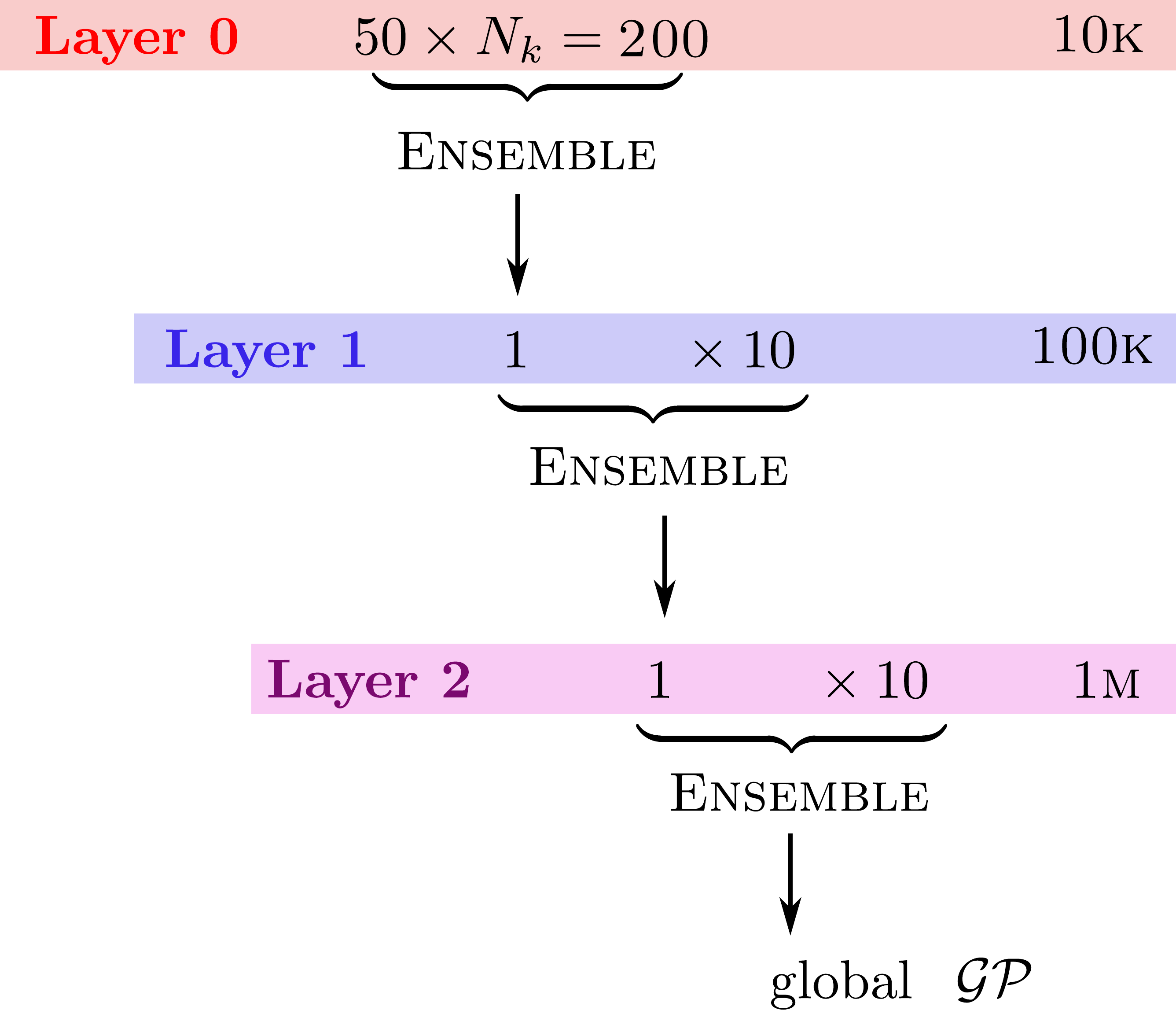}
	\caption{Graphical depiction of the \textit{pyramidal} structure for ensembles of ensemble GPs.}
	\label{fig:pyramid}
\end{wrapfigure}

\textbf{iii) Recyclable ensembles:} For simulating potential scenarios with at least $N=10^{6}$ input-output data points, we used the setting of the previous experiment, but with $K=5\cdot 10^{3}$ tasks of $N_k=800$ instead. However, as explained in the paper, its performance was hard to evaluate in the baseline methods, due to the problem of combining bad-fitted GP models. Then, based on the experiments of \citet{deisenroth2015distributed} and the idea of building ensembles of ensembles, we set a \textit{pyramidal} way for joining the distributed local GPs. It was formed by two \textit{layers}, that is, we joined ensembles twice as shown in the Figure \ref{fig:pyramid} of this appendix.

\textbf{iv) Solar physics dataset:} We used the solar physics dataset (\url{https://solarscience.msfc.nasa.gov/}) which consists of $N=3196$ samples. Each input-output data point corresponds to the monthly average estimate of the sunspot counting numbers from 1700 to 1995. The output targets $\yc$ were transformed to the real-valued domain via the mapping $\log(y_i + 1)$ to use a normal likelihood distribution. We also scaled the input area to the range $\xc \in [0, 100]$ and normalized the outputs to be zero-mean. The number of tasks was $K=50$ and $20\%$ of the data observations were reserved for test. The initial values of kernel and likelihood hyperparameters was $\{\ell = 0.2, \sigma^{2}_a = 1.0,  \sigma^{2}_n = 0.1\}$ where $\sigma^{2}_n$ is the initial likelihood variance, that we also learn. In this case, the setup of the VEM algorithm was \{$\textsc{ve}=20$, $\textsc{vm}=20$, $\eta_{m} = 10^{-5}$, $\eta_{L} = 10^{-8}$, $\eta_{\psi} = 10^{-10}$, $\eta_{Z} = 10^{-10}$\}. The number of global inducing-inputs used for the ensemble was $M=90$, whilst we used $M_k=6$ for each distributed approximation.

\textbf{v) Pixel-wise MNIST classification:} We took images of \textit{ones} and \textit{zeros} from the \textsc{MNIST} dataset. To simulate a pixel-wise unsupervised classification problem, true labels of images were ignored. Instead, we threshold the pixels to be greater or smaller than $0.5$, and labeled as $y_i=0$ or $y_i=1$. That is, we turned the grey-scaled values to a binary coding. Then, all pixels were described by a two-dimensional input in the range $[-1.0, 1.0]$, that indicates the coordinate of each output datum. In the case of the \textit{zero} image, we splitted the data in four areas, i.e.\ the four corners, as is shown in the subfigure (A) of Figure 4. Each one of the local tasks was initialized with an equally spaced grid of $M_k=16$ inducing-inputs. The ensemble GP required $M=25$ in the case of the number \textit{zero} and $M=16$ for the \textit{one}. The plotted curves correspond to the test GP predictive posterior at the probit levels $[0.2, 0.5, 0.8]$. The setup of the VEM algorithm was \{$\textsc{ve}=20$, $\textsc{vm}=10$, $\eta_{m} = 10^{-3}$, $\eta_{L} = 10^{-5}$, $\eta_{\psi} = 10^{-6}$, $\eta_{Z} = 10^{-5}$\}.

\textbf{vi) Compositional number:} As an illustrative experiment of the capabilities shown by the recyclable GP framework, we generated a number $\textit{eight}$ pixel-wise predictor whitout observing any picture. Instead, we used the $K=4$ distributed tasks of the previous experiment with the number \textit{zero}. We replicated the objects $\Ecal_k$ twice, so the final ensemble input list was $\{\Ecal_1, \dots, \Ecal_4, \Ecal_5, \dots, \Ecal_8\}$. The set of partitions $\{\Ecal_5, \dots, \Ecal_8\}$ was identical to the previous ones but we shifted the inducing-inputs $\bm{Z}_{k}$ by adding $1.2$ in the vertical axis. That is, with smaller distributed tasks of two number \textit{zeros}, we generated an ensemble of a number \textit{eight}. We remark that this experiment is purely illustrative to show the potential uses of the framework in compositional learning applications. The initial values of hyperparameters and the setup of the optimization algorithm was equivalent to the previous experiment.

\textbf{vii) Banana dataset:} The banana experiment is perhaps one of the most used datasets for testing GP classification models. We followed a similar strategy as the one used in the \textsc{MNIST} experiment. After removing the $33\%$ of samples for testing, we partitioned the input-area in four quadrants, i.e.\ as is shown in Figure 4. For each partition we set a grid of $M_k=9$ inducing-inputs and later, the maximum complexity of the global sparse model was set to $M=25$. The baseline GP classification method also used $M=25$ inducing-inputs and obtained an \textsc{nlpd} value of $7.29\pm7.85\times10^{-4}$ after ten trials with different initializations. Our method obtained a test \textsc{nlpd} of $7.21\pm 0.04$. As we mentioned in the main manuscript, the difference is understandable as the recyclable GP framework used a total amount of $4\times 16$ inducing-inputs, that capture more uncertainty than the $16$ of the baseline method. The setup of the VEM algorithm was \{$\textsc{ve}=20$, $\textsc{vm}=10$, $\eta_{m} = 10^{-3}$, $\eta_{L} = 10^{-5}$, $\eta_{\psi} = 10^{-6}$, $\eta_{Z} = 10^{-5}$\}.

\textbf{viii) London household data:} Based on the large scale experiments in \citet{hensman2013gaussian}, we obtained the register of properties sold in the Greater London county during the 2017 year (\url{https://www.gov.uk/government/collections/price-paid-data}). All addresses of household registers were translated to \textit{latitude-longitude} coordinates, that we used as the input data points. In our experiment, we selected two heterogeneous registers, one real-valued and the other binary. The real-valued output targets correspond to the log-price of the properties included in the registers. Moreover, the binary values make reference to the type of contract, $y_i = 1$ if it was a \textit{leasehold} and $y_i = 0$ if \textit{freehold}. Interestingly, we appreciated that both tasks share modes accross the input region, as they are correlated. That is, if there is more presence of some type of contract, it makes sense that the price increases or decreases accordingly. Therefore, after dividing the area of London in the four quadrants $\{Q1, Q2, Q3, Q4\}$ shown in the last Figure of the manuscript, we trained the $Q1$ exclusively with the regression data. Our assumption is that there exists an underlying function $f$ that is linked differently to the parameters depending if the problem is regression or classification. With this in mind, we trained the ensemble bound on the entire area of the city with two local GPs, one coming from regression in $Q1$ and the other from classification in $\{Q2,Q3,Q4\}$. To check if the error results showed an improvement in prediction, we compared the posterior GP prediction of the ensemble GP on $Q1$ with the local GP that did not observe any data on $Q1$. Results showed us, that even after not observing any binary data in $Q1$, the global GP performed better that the local GP with no regression information. The setup of the VEM algorithm was \{$\textsc{ve}=20$, $\textsc{vm}=10$, $\eta_{m} = 10^{-5}$, $\eta_{L} = 10^{-7}$, $\eta_{\psi} = 10^{-8}$, $\eta_{Z} = 10^{-7}$\} and we used $M=25$ inducing-inputs.

\subsubsection*{Appendix E.2. ~Performance metrics}

In our experiments, we used three metrics for evaluating the predictive performance of the global GP solutions: i) negative log-predictive density (\textsc{nlpd}), ii) root mean square error (\textsc{rmse}) and iii) mean absolute error (\textsc{mae}). Given a test input datum $\xc_t$ and $\{\hat{f}_t,\hat{y}_t\}$ being the predictive mean of the GP function and output prediction respectively, the metrics can be computed as
\begin{align*}
	\textsc{nlpd} &= - \sum_{t=1}^{N_t}\log p(y_t| \Dcal),\\
	\textsc{rmse} &= \sqrt{\frac{1}{N_t}\sum_{t=1}^{N_t}(\hat{f}_t - f_t)^{2}},\\
	\textsc{mae} &= \frac{1}{N_t}\sum_{t=1}^{N_t}\left|\hat{f}_t - f_t\right|,
\end{align*}
where $y_t$ and $f_t$ are the true output target and function values. $N_t$ is the number of test data points.

\subsubsection*{Appendix E.3. ~Optimization algorithms}

The following version of the variational expectation-maximization (VEM) algorithm is used both for the local and global inference of GPs. That is the reason why we do not include the sub-scripts $\{k,*\}$ in the parameter variables.

\begin{algorithm}[]
	\caption{--- \textsc{Variational EM for recyclable GPs}}
	\label{alg:vem}
	\begin{algorithmic}[1]
		\STATE Initialize $\bm{\psi}, \bm{\phi}$ and $\bm{Z}$\\
		\WHILE{\textbf{not} $\Lcal^{(t)}_{\mathcal{E}} \approx \Lcal^{(t-1)}_{\mathcal{E}}$}
		\STATE \# Variational Expectation (VE)
		\FOR{$j \in 1,\dots, \textsc{ve}$} 
		\STATE	update $\bm{\mu}_{(j)} \leftarrow \bm{\mu}_{(j-1)} + \eta_{\mu}\nabla_{\mu}\Lcal_{\mathcal{E}}$
		\STATE	update $\bm{L}_{(j)} \leftarrow \bm{L}_{(j-1)} + \eta_{L}\nabla_{L}\Lcal_{\mathcal{E}}$
		\ENDFOR
		\STATE \# Variational Maximization (VM)
		\STATE \# Hyperparameters
		\FOR{$j \in 1,\dots, \textsc{vm}$} 
		\STATE	update $\ell_{(j)} \leftarrow \ell_{(j-1)} + \eta_{\psi}\nabla_{\ell}\Lcal_{\mathcal{E}}$
		\STATE	update $\sigma_{a,(j)} \leftarrow \sigma_{a,(j-1)} + \eta_{\psi}\nabla_{\sigma_a}\Lcal_{\mathcal{E}}$
		\ENDFOR
		\STATE \# Inducing-inputs
		\FOR{$j \in 1,\dots, \textsc{vm}$} 
		\STATE	update $\bm{Z}_{(j)} \leftarrow \bm{Z}_{(j-1)} + \eta_{Z}\nabla_{Z}\Lcal_{\mathcal{E}}$
		\ENDFOR
		\ENDWHILE
		%\STATE {\bfseries input:} Observe $\Dcal^{(0)}_\text{new}$ 
		%\STATE	Maximise 
		%\FOR{$t \in 1,\dots, T$}
		%\STATE	Update 
		%\STATE Choose initial 
		%\STATE Compute continual GP prior 
		%\STATE {\bfseries input:} Observe $\Dcal^{(t)}_\text{new}$ 
		%\STATE	Maximise 
		%\ENDFOR
	\end{algorithmic}
\end{algorithm}

For the distributed GP regression models needed for the baseline methods, we used the LBFGS optimization algorithm with a learning rate $\eta=10^{-2}$. We set a default maximum of $50$ iterations.

\end{document}

%% file: definitions.tex
\def\K{{\mathbf K}}

\def\L{{\mathbf L}}

\def\f{{\mathbf f}}

\def\0{{\mathbf 0}}

\newcommand{\Ecal}{\mathcal{E}}

\newcommand{\Ocal}{\mathcal{O}}
\newcommand{\Ncal}{\mathcal{N}}
\newcommand{\Dcal}{\mathcal{D}}

\newcommand{\Lcal}{\mathcal{L}}

\newcommand{\fc}{\bm{f}}
\newcommand{\uc}{\bm{u}}
\newcommand{\xc}{\bm{x}}
\newcommand{\yc}{\bm{y}}
\newcommand{\zc}{\bm{z}}

\newcommand{\finf}{f_{\infty}}
\newcommand{\up}{\bm{u_*}}
\newcommand{\uk}{\bm{u}_k}

\newcommand{\qt}{q_{\mathcal{C}}(\uk)}

\newcommand{\Ebb}{\mathbb{E}}
\newcommand{\muk}{\bm{\mu}_k}

\newcommand{\Sk}{\bm{S}_k}
\newcommand{\Kk}{\K_k}

\newcommand{\Sp}{\bm{S_*}}
\newcommand{\mup}{\bm{\mu_*}}

\newcommand{\Kpp}{\K_{\bm{**}}}
\newcommand{\Kpk}{\K_{\bm{*}k}}

\newcommand{\cmark}{\ding{51}}%
\newcommand{\xmark}{\ding{55}}%